\documentclass[10pt,journal,compsoc]{IEEEtran}

\ifCLASSOPTIONcompsoc
  \usepackage[nocompress]{cite}
\else
  \usepackage{cite}
\fi

\usepackage[dvips]{graphicx}
 \usepackage{amsmath}
\usepackage{amsmath,algorithm}
\usepackage{amsthm}
\usepackage{subcaption}
\usepackage{bm}

\usepackage[noend]{algpseudocode}
\usepackage{setspace}
\usepackage{color}
\usepackage{graphics}
\usepackage{tikz}
\usetikzlibrary{arrows}
\usepackage{mathrsfs}
\usepackage{amssymb}
\usepackage{dsfont}

\usepackage{cases}
\usepackage{relsize}
\usepackage{tablefootnote}
\usepackage{threeparttable}
\usepackage{booktabs}
\usepackage{makecell}
\usepackage[markup=hmarkupi]{changes}

\usetikzlibrary{arrows}

\algnewcommand\And{\textbf{and}}
\algnewcommand\Or{\textbf{or}}

\makeatletter
\def\endthebibliography{%
  \def\@noitemerr{\@latex@warning{Empty `thebibliography' environment}}%
  \endlist
}
\makeatother

\DeclareMathOperator{\E}{\mathbf{E}}

\newcommand{\avg}[1]{\left\langle #1\right\rangle}

\newcommand{\M}[1]{{\bf #1}}
\newcommand{\V}[1]{{\bf #1}}

\newcommand\given[1][]{\:#1\vert\:}

\begin{document}

\title{Data-Driven Interaction Analysis of Line Failure Cascading in Power Grid Networks}

\author{Abdorasoul~Ghasemi and~Holger Kantz
\IEEEcompsocitemizethanks{\IEEEcompsocthanksitem A. Ghasemi is with the Department
of Computer Engineering, K. N. Toosi University of Technology, Tehran, Iran, E-mail: arghasemi@kntu.ac.ir.
\IEEEcompsocthanksitem Holger Kantz is with the Max Planck Institute for Physics of Complex Systems, Nöthnitzer Str. 38, 01187 Dresden, Germany}
\thanks{Manuscript received January xx, 2021; revised January\ xx, 2021.}}


\IEEEtitleabstractindextext{%
\begin{abstract}
We use machine learning tools to model the line interaction of failure cascading in power grid networks. We first collect data sets of simulated trajectories of possible consecutive line failure following an initial random failure and considering actual constraints in a model power network until the system settles at a steady state. We use weighted $l_1$-regularized logistic regression-based models to find static and dynamic models that capture pairwise and latent higher-order lines' failure interactions using pairwise statistical data. The static model captures the failures' interactions near the steady states of the network, and the dynamic model captures the failure unfolding in a time series of consecutive network states. We test models over independent trajectories of failure unfolding in the network to evaluate their failure predictive power. We observe asymmetric, strongly positive, and negative interactions between different lines' states in the network. We use the static interaction model to estimate the distribution of cascade size and identify groups of lines that tend to fail together, and compare against the data. The dynamic interaction model successfully predicts the network state for long-lasting failure propagation trajectories after an initial failure.     
\end{abstract}

}

\maketitle

\IEEEdisplaynontitleabstractindextext

\IEEEpeerreviewmaketitle

\section{Introduction}\label{sec: intro}

Network robustness is an emerging need for networked systems and refers to the system's ability to operate effectively after possible component-level disruptions, or environment changes \cite{savla2019network}. Failure cascading process is a high-risk event in networked systems in which the overall cost, e.g., the number of shutdown users in the power grid, increases in the same order as the probability of the event decreases. In networked systems, the direct and indirect interactions between the system components induce correlations and may amplify or attenuate the initial disturbance. The amplification \cite{moore1956reliable} or attenuation \cite{ronellenfitsch2018optimal} effects of network structure after especially correlated fluctuations reflect the underlying interplay between the structure and dynamics of the complex networked systems.

Network science helps to understand the system's robustness by studying the direct and indirect interactions among the system's elements after a perturbation. The sparse and hierarchical structure of natural biological networks is related to their robustness against fluctuations taking into account the cost of the indirect interactions in  \cite{ronellenfitsch2018optimal} and \cite{leclerc2008survival}. Consistently, results of \cite{yaziciouglu2016resilience} show that adding a new link or increasing the capacity of a link may have adverse effects and decreases the resilience of networks with locally routed flows. The results of these studies suggest that beyond the pairwise interaction analysis, we need new tools to capture the non-trivial indirect higher-order interactions for analyzing the robustness of networked systems.

In power networks, lines' failure cascading are correlated in a non-trivial pattern, rarely leading to large blackouts according to the historical data \cite{carreras2004evidence}. The origins of cascading process in power networks are related to the self-organized criticality phenomenon in complex systems in \cite{carreras2004complex}  \cite{dobson2007complex} and more recently is linked to the power-law nature of city inhabitants \cite{nesti2020emergence}. Some other studies, instead of finding what gives rise to the phenomenon, focus on finding how the cascade process relates to the network's structure and how it unfolds in the network in a deterministic \cite{guo2017monotonicity} or stochastic manner \cite{yang2017vulnerability}. These studies link the failure unfolding process to the pairwise line interaction. The pairwise line interaction refers to the mutual impact that a pair of lines has on each other after a failure of one of them.

In \cite{guo2017monotonicity, guo2020localization}, the authors use the deterministic pairwise line outage redistribution factors (LODFs) and matrix-tree theorem to analyze how failure propagates through spanning forests in the network graph if the network remains connected. In many failure cascading scenarios, however, the network partitions into some islands. On the other hand, data-driven approaches rely on analyzing pairwise line interactions statistics after different initial failure scenarios. Reference \cite{yang2017vulnerability} suggests a two-stage algorithm to identify the co-susceptible line groups using the pairwise failure correlation matrix in a stochastic manner. The first stage determines the significantly correlated pairs, and in the second stage, an agglomerating algorithm finds cliques with enough correlation as co-susceptible groups. The model based on extracted co-susceptible groups is then used to estimate the cascade size statistics as a complex response and compared with the simulated data. As we shall discuss, pairwise correlations do not capture some crucial interactions. In \cite{hines2016cascading}, the authors use the pairwise line failure statistics in generations, i.e., consecutive cascade unfolding time-steps cascade unfolding, to find the influence graph. Assuming that the number of total outages propagated by each line failure is Poisson, the authors find the probability of pairwise line outage propagation. The inferred influence graph is then used to predict the cascade size and compared against the simulated data if failures propagate locally over this graph. 

Although finding the pairwise statistics is straightforward and computationally tractable even for large networks, they are not sufficient per se if higher-order interactions exist. Despite the pairwise interaction, in higher-order interactions, the simultaneous states of more than two lines are involved in determining the system dynamics. Higher-order interactions may substantially affect the dynamics of complex networked systems \cite{battiston2021physics}. The failure cascading process in power grid networks involves higher-order interactions, as we discussed in more detail in subsection ~\ref{sec: higher-order}. However, collecting data for possible higher-order interactions is not straightforward, if even possible, due to the explosive number of possible combinations. Therefore, there is an interest in finding the possible higher-order interaction using the ordinary pairwise statistics.

The authors of \cite{schneidman2006weak} show that maximum entropy statistical models can successfully capture the higher-order interaction of neural activity dynamics using the ordinary pairwise correlation data. Next, Ref. \cite{aurell2012inverse} shows that the Pseudo-likelihood and approximate maximum entropy statistical model can successfully recover the interaction topology even from a limited amount of data. These results motivate us to investigate the sufficiency and learn statistical models from pairwise statistics that capture the underlying higher-order interactions topology of line failure cascading in power networks.

This paper considers the inverse problem of learning the interaction graph from the pairwise statistics collected from simulated data of line failures in the steady states and over time. We first discuss that the failure cascading process in power grid networks involves higher-order interactions overlooked by observing the pairwise correlation data. Next, we aim to learn statistical models that capture the latent higher-order lines failure interactions. The models use ordinary pairwise statistics data to successfully predict complex system responses like the cascade size statistics and consecutive network state. We find static and dynamic interaction graphs. The static interaction graph helps us to estimate the cascade size distribution and identify lines that fail together. On the other hand, the time series analysis helps find how the failure unfolds in the network.    


The rest of this paper is organized as follows. In Section~\ref{sec: models}, the system model and physics of the flow distribution in the power network are discussed, and the process of collecting the data is explained. The possible Section~\ref{sec: interactions} extends the pairwise interaction to higher-order interactions and their importance by providing illustrative examples. Here, we also discuss the sufficiency and statistical models which can capture the higher-order interaction using pairwise statistics. In Section~\ref{sec: ss-modeling}, we explain the learning process of inferring the line interactions in the network steady-state and using the learned model to infer the co-susceptible group of lines. In Section~\ref{sec: ts-modeling} we discuss the learning of interaction matrix that encodes how the cascade unfolds in the network before concluding in Section~\ref{sec: conclusion}.

\section{Models and data set preparation} \label{sec: models}
\subsection{System model} \label{sec: system_model}
Consider a power grid network with $\mathcal{N}=\{1,\ldots,n\}$ buses or nodes and $\mathcal{E} \subset \mathcal{N} \times \mathcal{N}, |\mathcal{E}|=L$,  transmission lines or edges with the corresponding graph $\mathcal{G}~=~(\mathcal{N},\mathcal{E})$. In the normal operation, the network facilitates the electricity flow distribution from generator buses to load buses meeting the underlying system's physics (Ohm's rule, flow conservation rule, and power balance) and its constraints, i.e., the maximum generation power of generators and the maximum capacity of lines.   

Ignoring the lines' resistances, the susceptance of line $e=(i,j) \in \mathcal{E}$ between bus $i$ and $j$ is given by $b_{ij} ~=~ \frac{1}{x_{ij}}$ where $x_{ij}$ is the line's reactance. Let $\M B_{L \times L}~=~\textrm{diag}(b_e: e\in \mathcal{E})$  and $\M C_{n \times L}$ denote, respectively, the susceptance and the node-link incidence matrix of $\mathcal{G}$ assuming an arbitrary orientation for each link. In this paper, all matrices and vectors are, respectively, denoted by bold uppercase and bold lowercase letters. The power injection or demand at bus $i$ is $p_i$ and $\V{p}_{n \times 1}=(p_1, \ldots, p_n)$ is the corresponding vector. $f_{e}$ is the flow on link $e$ and $\V{f}_{L \times 1} = (f_1, \ldots, f_L)$ is the flow vector of the network. Assume that the voltage magnitude of all buses is normalized to 1 and the unknown voltage phase of bus $i$ is denoted by $\theta_i$. In the linear model, applying Ohm's law for link $e=(i,j)$ we have $f_{e} = (\theta_i - \theta_j) b_{ij}$, which in  the matrix form reads as $\V f (t) = \M B(t) \M C(t)^{T} \boldsymbol \theta(t)$. The flow conservation law at each bus meets, $\M C(t) \V f(t) = \V p (t)$. Ohm's law and flow conservation, along with the power balance constraint $\V 1^{T} \V p (t) = 0 $, ends up to finding $n-1$ unknown voltage phases assuming the voltage phase of the slack bus generator as zero. The power of the slack bus adjusts to meet the small fluctuations in the power supply-demand balance in the network. Specifically, let $\M L(t) = \M C(t) \M B(t) \M C^{T}(t)$ denote the Laplacian matrix of the $\mathcal{G}$, i.e., $L_{ij} = -b_{ij}$ if there is a link between $i$ and $j$ and $L_{ii}=\sum_{j}b_{ij}$.  The voltage phases are then given by $\boldsymbol \theta(t) = \M L^{\dagger}(t)  \V p (t)$ where $\M L^{\dagger}$ is the Moore-Penrose inverse of $\M L$. Finally, using Ohm's law the flow of each line reads as $\V f (t) = \M B(t) \M C(t)^{T} \M L^{\dagger}(t)  \V p (t)$.

Each generator has a capacity above it will shut down. Also, there is a capacity for line $e$, $c_{e}$, and the line fails if its flow exceeds its capacity. Therefore, the steady-state lines' flows are the solutions of the above linear model subject to many physical constraints. The network is subject to line failure perturbations in time, e.g., due to lightning or malfunctioning of relays. After the initial failure, the flows are redistributed. This may lead to subsequent line failures, power imbalance, and even partitioning the network before the network settles in a new steady state. This linear flow distribution and redistribution model in the power grid captures essential features of the cascade process like a non-additive response, non-local propagation, and disproportional impact \cite{crucitti2004model} and is used in other works\cite{carreras2002critical}, \cite{yang2017vulnerability}.

\subsection{Data set preparation}\label{sec: data-set}
We develop a simulator to collect a data set of failure cascading trajectories for given network topology, power generation and demands at buses, the maximum power of generators, and the capacity of lines.  

The initial flow of each line is computed using the flow distribution model, assuming all lines are working properly. At each run, the process starts with randomly removing a small random subset of lines in which each line is removed independently with probability $p_f$. We set  $p_f = \frac{2.5}{L}$ in our data collection phase. Next, the new line flows are recomputed, and if a line's flow exceeds its corresponding capacity, that line fails as well, which may trigger other consecutive failures. We record the failed lines at each time step until the network settles at a steady state, see Fig.~\ref{figs: GTOC}. The network may disconnect due to failures and decompose into components. Therefore, the power balancing of the network or its components may be destroyed. We adopt the power re-balancing strategy explained in \cite{hines2011topological}. In this strategy, the small power imbalance is compensated by ramping up or ramping down the power generation at generators. Beyond that, we use generator tripping and load shedding with the priority of small generators or loads. We simulate and collect $M$ trajectories of failure cascading on the IEEE-118-Bus network. The required data, including the network connectivity, the lines' capacities, and the maximum generators' powers, are available in  \cite{IEEE-118}. The basic statistics of IEEE-118-Bus network are $N=118$, $L=179$; mean degree $\avg k = 3.034$; clustering coefficient $C = 0.136 $.

We perform our experiments on two data sets. The first data set $D_1$ consists of $M \approx 52000$ unique trajectories with random initial failure scenarios. Due to the available redundancies, many initial failures do not propagate. In this data set, 46\% of the initial failures lead to at least one consecutive failure, while the remaining 54\% do not propagate.  This data set is used to infer the interactions in the normal operation of the network.  Data set $D_2$ consists of about $M \approx 38000$ trajectories in which all of the initial failures propagate at least one step. The interaction matrix from this data highlights the indirect interactions in the cascading scenarios.

In the following the state of line $i$, is denoted by $s_i = \pm 1$, where $s_i=+1$ indicates that the line fails. The state of network is completely determined by $\V s(t)=(s_1,\ldots,s_L)$. We measure the cascade size, $Z$, in terms of the number of failed lines, $Z = \sum\limits_{i=1}^{L}\frac{(1+s_i)}{2}$. Note that, although the details of simulations like the power balancing strategy affect the collected data sets, the main interesting feature of observing heavy tail distribution in the cascade size remains unchanged. We are interested in exploiting these data to learn statistical models which encode the lines' interactions and use them to infer lines that fail together, the influential lines, and how the cascade unfolds in time.

\section{Pairwise and higher-order interactions} \label{sec: interactions}
We first present the pairwise line failure interaction in the power network. We review previous deterministic and statistical results that show the relationship between the pairwise interaction's absolute value and the physical adjacency of the corresponding lines in the power network. We use this prior knowledge in our learning scheme in Section \ref{sec: ss-modeling}. Next, we discuss possible higher-order interactions which might be overlooked by observing the pairwise correlations directly. Finally, we discuss the statistical models that we use to capture the higher-order interactions using pairwise data. 


\subsection{Pairwise interactions} \label{sec: pair-wise}

For a given pair of lines, the (asymmetric) pairwise interaction shows to what extent one line's failure may lead to consecutive overload or failure of the other line. Let $(a,b)$ denote the line between nodes $a$ and $b$ and consider the pair $e=(a,b)$ and $\hat{e}=(c,d)$. Assume  $e$ fails. The line outage redistribution factor (LODF), $K_{e\hat{e}}$, is the ratio of flow changes on line $\hat{e}$ to the initial flow on line $e$ before it was failed provided that the network remains connected. $K_{e\hat{e}}$ is independent of the power injection or demand vector $\V p$ and only depends on the underlying weighted graph and can be efficiently computed deterministically \cite{guo2017monotonicity}. 
Specifically, $K_{e\hat{e}}$ depends on the weight of certain spanning forests in graph $\mathcal{G}$ . In particular, if $e$ and $\hat{e}$ are connected to a common bus we have $K_{e\hat{e}} > 0$. That is, the proximity in the physical network usually implies interactions as we expected. 
Alternatively, one could find the pairwise line failure correlations using a reasonable amount of recorded data or simulation. The statistical failure analysis results show that the farther the distance between the lines, the less strong the interaction value is \cite{witthaut2015nonlocal}. Note that we observe physically far distance but strong interacting line pairs as well. 

We use this prior knowledge to adjust the regularization (penalization) factor in the process of learning the interaction structure between lines. We adopt the edge distance, $d_{e,\hat{e}}$, which was introduced in \cite{witthaut2015nonlocal} to investigate the non-local effect of failure cascading. Let $d_{x,y}$ denote the shortest path length between nodes $x$ and $y$ in $\mathcal{G}$. We have
$d_{e,\hat{e}} = \min_{x \in \{a,b\}, y \in \{c,d\}} d_{x,y} + 1$. Note that if $e$ and $\hat{e}$ are connected to a common bus $d_{e,\hat{e}} =1$.

\subsection{Higher-order interactions} \label{sec: higher-order}

Pair-wise statistics of lines' failures are not sufficient per se if the cascade process involves higher-order interactions. Higher-order interaction refers to a group of more than two lines whose simultaneous states affect system dynamics.  The existence of higher-order interactions in failure cascading is also mentioned in previous studies. The authors of \cite{yang2017vulnerability} believed that the discrepancy between their expected model results and data at higher loads are related to higher-order correlations, which are not captured by the correlation matrix. Also, \cite{witthaut2015nonlocal} shows that by intentional removal of a specific link, we can mitigate the cascading effects, which shows that there is non-trivial indirect interaction between the failures of lines' groups.

\begin{figure}
	\centering
	\begin{subfigure}[b]{0.15\textwidth}
		\centering
		\includegraphics[width=\textwidth]{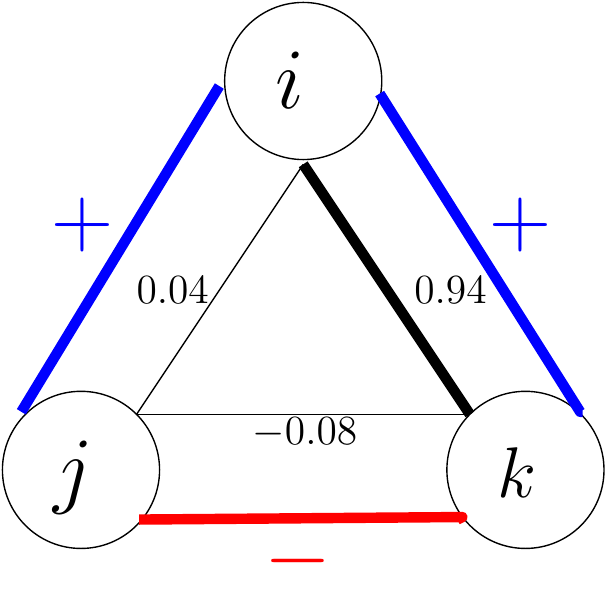}
		\caption{}    
		\label{figs: ts-fitness-D1}
	\end{subfigure}\hfill
	\begin{subfigure}[b]{0.3\textwidth}  
		\centering 
		\includegraphics[width=\textwidth]{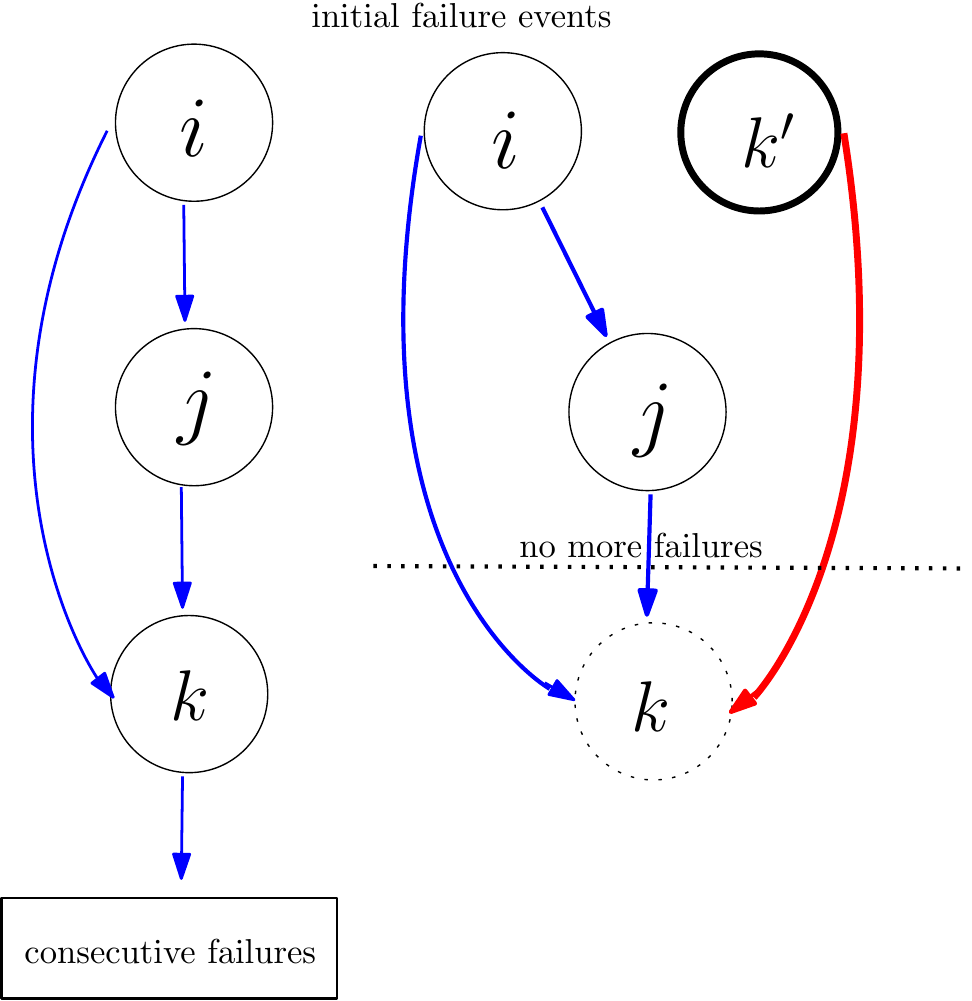}
		\caption{}    
		\label{figs: ts-fitness-D2}
	\end{subfigure}
	
	\caption{(a) The three-way interactions among three selected lines are shown as a frustrated triplet. The pairwise Pearson correlation coefficients are shown in the inner triangle. We show positive interactions in blue and negative interactions in red. Due to the negative interaction between $k$ and $j$, we do not observe a significant correlation between the failures of $i$ and $j$. (b) Compared with the initial failure of $i$ (left), the simultaneous failure of $i$ and $k^\prime$ (right) avoids subsequent failure of $k$ and the following cascading due to negative interaction between the failure of $k$ and $k^\prime$.}\label{figs: frust-triplet}
\end{figure}

We provide two illustrative examples to explain these indirect interactions and their importance in our subsequent inference and network dynamics. The first one is an example of third-order interactions between a selected group of three lines which are overlooked by direct observing pairwise correlations. The second example shows that we can mitigate the cascade effect by intentionally shutting down a line to exploit the possible negative interaction between a specific line group. We use the collected data for failure cascading in power networks in data set D2.

Let $i$, $j$, and $k$ denote, receptively, lines (3,5), (7,12), and (5,6) in the IEEE-118 network as shown in Fig.~\ref{figs: 118-interaction}. Assume $C_{xy}$ denote the Pearson correlation coefficient between $x$ and $y$. Using data set $D_2$ we have $C_{ik} = 0.94$, $C_{ij} = 0.04$, and $C_{kj} = -0.08$. See Fig. \ref{figs: frust-triplet}(a). 
Therefore, pairwise correlations show that the failure of lines $i$ and $k$ are strongly correlated, and there is no significant correlation between the failures of $i$ and $j$. Now let $C_{x,y | z}$ denote the correlation between lines $x$ and $y$ given the state of line $z$. We have $C_{i,j| k=-1} = 0.43$ and $C_{i,j | k=+1} = -0.005$. If line $k$ does not fail, then there is a significant correlation between the failures of $i$ and $j$, while if line $k$ fails, there is not. Here, we observe statistically significant three-way interaction, which is overlooked by pairwise interactions. 

Next, let $J_{xy}$ denote the interaction value for lines $x$ and $y$ predicted by the learned statistical models in Section \ref{sec: ss-modeling}. The learned model predicts strong positive bi-directional  interaction between $i$ and $j$ and so do $k$ and $i$, i.e., $J_{ij}, J_{ji} \gg 0 $ and $J_{ki}, J_{ik} \gg 0$. However, it predicts strong negative bi-directional interaction between $j$ and $k$, i.e., $J_{jk},  J_{kj} \ll 0$. We find that the weak correlation between the failure of $i$ and $j$ roots at the strong negative interaction between the failure of $j$ and $k$. In scenarios in which $i$ and $k$ fail, $j$ did not fail, consistent with the data. These third-order interactions, named the frustrated triplets, are not considered by simply looking at the pairwise correlations. This example shows that we can not rely on 
the naive pairwise correlation coefficient, for example, to infer the groups of lines that fail together as some strong interaction might be overlooked.   

Fig.~\ref{figs: frust-triplet}(b) shows another example of the impact of finding the higher-order interaction in the cascade dynamics. In this example we have $i=(26,25), j=(30,38), k=(17,18)$ and $k^\prime=(18,19)$. Here we observe how the strong negative interaction between the failure of line $k$ and $k^\prime$ can mitigate the cascade effect. The initial failure event of the line $i$ leads to overload and failure of the line $j$. Next line $k$ fails, and we observe a series of consecutive line failures that fails 12 other lines. However, if in the initial failure event, $i$ and $k^\prime$ fail simultaneously, we observe that $j$ fails and the process stops. Our temporal interaction analysis in Section~\ref{sec: ts-modeling} shows that there is a strong negative interaction between the failures of $k$ and $k^\prime$; suggesting that we can prevent the failure of $k$ and its subsequent failures by intentionally failing $k$ in this scenario. 

\subsection{Class of pairwise statistical models} \label{sec: statistical-model}
The higher-order interaction impacts the failure cascading process in power grid networks. However, collecting the required data for these latent interactions is computationally inefficient considering the explosive number of possible combinations. The class of pairwise statistical models is of particular interest as they may find the latent higher-order interactions using the ordinary pairwise data, which can be collected conveniently from a moderate amount of data. Pairwise models assume that the response of each element in the networked system results from its pairwise interactions with some not-necessarily local elements. The efficiency of the pairwise statistical model to capture higher-order correlations was first observed in the study of strongly correlated network states of neural activity dynamics in \cite{schneidman2006weak}.

For binary variables, Ising and kinetic Ising models are general graphical models in this class for stationary statistics \cite{lokhov2018optimal} and are widely used in inverse problems using data, see \cite{nguyen2017inverse} for a survey. The inverse Ising model is used when the underlying interaction matrix is symmetric and the detailed-balanced equations between the network states are held. Consequently, we have a probability distribution, e.g., Gibbs maximum entropy, which assigns a probability to each network state based on its energy. However, in the kinetic Ising model with asymmetric interactions, the underlying probability distribution for steady states is not known in general.

The sufficiency of pairwise interactions to capture complex interactions in a non-perturbative regime is related to the sufficient constrained network states in \cite{merchan2016sufficiency}. In networked systems, by engineering or evolution, we observe many degrees of freedom and many constraints. Take the considered power network in the linear model as an engineered system. The lines' flows and generators' powers are degrees of freedom to ensure proper operation. However, the system operation is subject to local and global constraints. Local constraints include the flow capacity of each line and the flow conservation rule at each node. The maximum power capacity of generators and power balance are two global constraints. The outcome of these constraints is the emergence of effective pairwise interactions, which couples system variables pairwise. These non-trivial pairwise interactions then explain the higher-order interactions. The effect of constraint density on the solution state space of random satisfiability problem (SAT) is studied in the theory of computation \cite{mezard2009information}. The high-density constraints lead to network state clusters or spin configurations that satisfy all constraints and can be explained by pairwise models.  

We use the machine learning techniques and prior knowledge of interactions' strength to find pairwise statistical models that capture the higher-order interactions of failure cascading and use these models for inference.

\begin{figure*}[ht!]
	\includegraphics[width=0.95\textwidth]{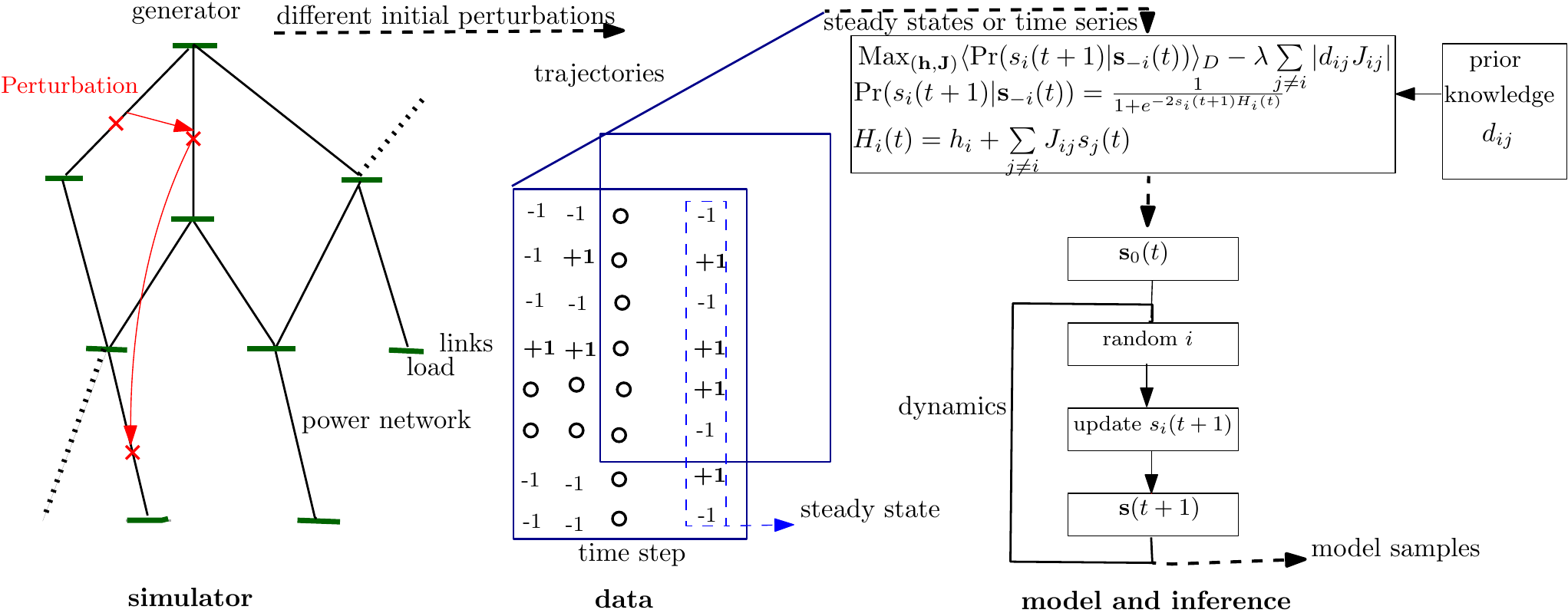}
	\caption{The overall flow diagram of a statistical learning-based approach for interaction modeling and inference of failure cascading in power grid networks}\label{figs: GTOC}
\end{figure*}

\section{Static interaction learning}\label{sec: ss-modeling}

Fig.~\ref{figs: GTOC} shows the diagram of the learning and inference procedure. We first consider a scenario in which we are interested in finding the static interaction graph, i.e., the relationship between a pair of lines' states at steady states. Note that according to the nature of the power networks, the desired interaction matrix is not symmetric in general. Consider lines $e$ and $\hat{e} $ which, respectively, connect a generator and a load to the network in a nearby neighborhood. The network is subject to tight constraints after $e$ fails, which probably leads to $\hat{e}$ failure. The failure of $\hat{e}$, on the other hand, makes the constraints lose and provide more slack power for the network. 
We expect to observe a collection of steady-state configurations in which all system constraints are met. The interaction graph at a steady state helps us understand which links tend to fail together and find co-susceptible groups. 

\subsection{Logistic regression model}
Let us single out link $i$ and assume that we have other links' states at time $t$ denoted by $\V s_{-i}(t)$. We can find $(h_i, \{J_{ij}, j \neq i\})$ such that the probability that link $i$ at $t+1$ is at proper state consistent with the data (constraints) is maximized where $J_{ij}$ is the influence of line $j$ on line $i$ and $h_i$ is a local factor. Specifically, let the state of link $i$ be related to other links' states according to 

\begin{align}\label{equ: ss-prob}
	\Pr(s_i(t+1) | \V s_{-i}(t)) & = \frac{1}{2}\left[1+s_i(t+1)\tanh(H_i(t))\right] \\ \nonumber &= \frac{1}{1 + e^{-2s_i(t+1) H_i(t)}}\, , \\
	H_i(t) &= h_i + \sum_{j \neq i} J_{ij} s_j(t).  \nonumber
\end{align}

Equ.(\ref{equ: ss-prob}) is a logistic regression estimator for $s_i$ conditioned on other links' states.
We should find $(h_i, \V J_i)$ by maximizing the log-likelihood function of observing $M$ independent $s_i(t+1)$ given $\V s_{-i}(t)$ over the data by $(h_i^*, \V J_i^*) = \textrm{argmax}_{(\V h_i, \V J_i)} \mathcal{L}_D(h_i, \V J_i)$ where
\begin{align}\label{equ: log-likelihood}
	\mathcal{L}_D(h_i, \V J_i) &= \frac{1}{M}\ln \prod_{m=1}^{M} \Pr(s_i(t+1) | \V s_{-i}(t)) \nonumber \\  &=  \avg{\ln\frac{1}{1 + e ^ {-2s_i(t+1) (h_i + \sum_{j \neq i} J_{ij}s_j)}}}_D.
\end{align}
$\V J_i$ is the $i$th row of interaction matrix and $\avg{f(\V s)}_D = \frac{1}{M} \sum_{m=1}^{M}f(\V s^{(m)})$ with data set
$D = \{\V s^1,\ldots,\V s^M\}$.

In practice, however, link $i$ does not effectively interact with all other links, and we are interested in finding a sparse solution in which the state of each link is presented in terms of explainable interactions that the physics of the problem dictates. In the $l_1$-regularized learning technique, to avoid finding spurious meaningless interactions, the penalizing term is added to the objective function of \eqref{equ: log-likelihood} considering the prior knowledge of the interactions. This penalizing term leads to set un-explainable interactions to zero. 

Let $\partial_i$ denote the neighbors of link $i$, i.e., the set of other lines with them $i$ has effective interaction. In \cite{lokhov2018optimal} the authors show that reconstruction of the interaction structure and strength is possible with a two-stage algorithm. In the first stage, we find the underlying graphical model by ruling out the weak interactions and finding the explanatory neighbor variables, $\partial_i, \forall i$. In this regard, we first solve $L$ independent optimization problems as 
\begin{align}
	(h_i^0, \V J_i^0) = \textrm{argmax}_{(\V h_i, \V J_i)} \mathcal{L}_D(h_i, \V J_i) - \lambda \sum_{j \neq i}|d_{ij} J_{ij}|,\label{equ: RPLE} 
\end{align}
where $\lambda$ is a regularization parameter and $d_{ij}$ is the distance between line $i$ and $j$ according to definition in subsection \ref{sec: pair-wise}. Here, we use the prior knowledge that the physically adjacent lines show greater interaction absolute value and hence less penalize the corresponding interaction in the optimization objective. Then all weak interactions with $-\delta_m < J_{ij} < \delta_p$ are set to zero, where $\delta_m, \delta_p > 0$ are proper thresholds.

In the second stage, having the interaction structure, we find the interaction strength $(h_i^*, \V J_i^*)$  by solving  \eqref{equ: RPLE} again with $\lambda = 0$. Note that we may end up with weak but important coupling at the end of the procedure.

Choosing appropriate $\lambda$ is related to the graphical model reconstruction problem and should be tuned for the inference problem. Assuming no other prior information, this parameter is related to the number of samples $M$, number of variables, $L$, and the accepted error in interaction graph reconstruction $\epsilon$, by $\lambda \propto \sqrt{\ln(L^2/\epsilon)/M}$ \cite{lokhov2018optimal}. $\delta_p$ and $\delta_m$ are then selected by inspecting the histogram of $\M J_i$ values near zero and identifying the gaps in the density of interaction strengths.

Note that by proper selection of $\lambda, \delta_p, \delta_m$ we can trade off the goodness of fit to data for the model complexity or finding a sparse interaction matrix. Also, the $l_1$-regularized logistic regression in \eqref{equ: RPLE}, is the conditional maximum entropy inference of $s_i(t+1)$ given $\V s_i(t)$, and benefits from the learning guarantees of this model \cite{mohri2018foundations}.

Computing the derivative of $\mathcal{L}_D(h_i, \V J_i)$ with respect to $h_i$ and $J_{ij}$, at the optimal point, we have 
\begin{align}\label{equ: ss-derivative}
	\avg {s_i} _D \approx \avg {\tanh (h_i^* + \sum_{k \in \partial_i} J_{ik}^{*} s_k)} _ D\\ \nonumber
	\avg {s_i s_j} _D \approx \avg {s_j \tanh (h_i^* + \sum_{k \in \partial_i} J_{ik}^{*} s_k)} _ D
\end{align}
which can be used as a measure of goodness of fit.

Learning $(\V h^*, \M J^*)$, we can use a dynamics which updates one link (spin) at each time step according to \eqref{equ: ss-prob} to find steady states. The Glauber dynamics is widely used in statistical physics for describing equilibrium and non-equilibrium Ising models as well as damage spreading modeling. The Glauber process starts with a random initial spin configuration. Next, at each time step one spin is selected randomly, say $i$, and updated, i.e., $s_i(t+1)$ takes value one with probability $\Pr(s_i(t+1)=1 | \V s_{-i}(t))= \frac{1}{1+e^{-2\left(h_i + \sum_{j \in \partial _i}J_{ij}s_j(t)\right)}}$. The Glauber dynamics should successfully reconstruct the network steady states if the underlying interaction matrix is leaned.    

\subsection{Interactions at steady states}
Since multiple initial failures may lead to the same steady state we first remove the final duplicate states in each data set. In the learning procedure, we use $\lambda_1 = 0.0001$ and $\lambda_2=0.0005$ for data sets $D1$ and $D2$. Also, we set $\delta_m=\delta_p=0.1$ to learn $(\V h_i, \M J_i)$ for all $i$. The maximum edge distance for the IEEE-118 network is 15.  

The optimization problem in \eqref{equ: RPLE} is convex and hence has a unique global optimum. However, the objective function is not differentiable if $\lambda \neq 0$. Therefore, in the first stage of the algorithm, we use proximal gradient descent, which shrinks the non-explanatory variable to zero in the projection step to find $(h_i^0, \V J_i^0)$ for each $i$.    

Using the selected parameters, we find sparse interaction matrices. The ratios of non-zero elements in $\M J_1^*$ and $\M J_2^*$  to all possible $L(L-1)$ interactions are $6.5\%$ and $5.8\%$.  

\begin{figure}
	\begin{subfigure}[b]{0.25\textwidth}
		\centering
		\includegraphics[width=.85\linewidth]{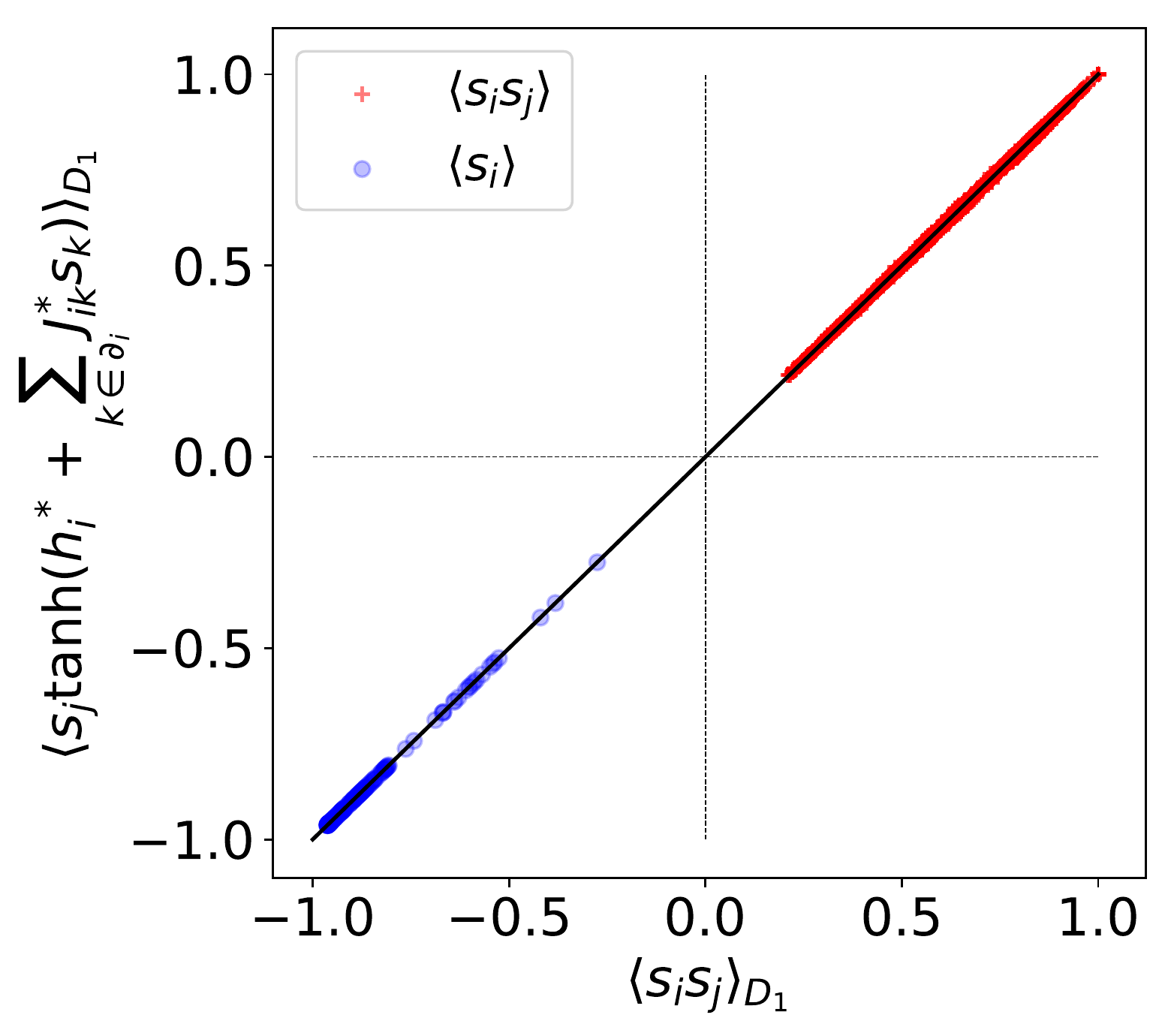}
		\caption{}
		\label{figs: ss-goodness-D1}
	\end{subfigure}%
	\begin{subfigure}[b]{0.25\textwidth}  
		\centering 
		\includegraphics[width=.85\linewidth]{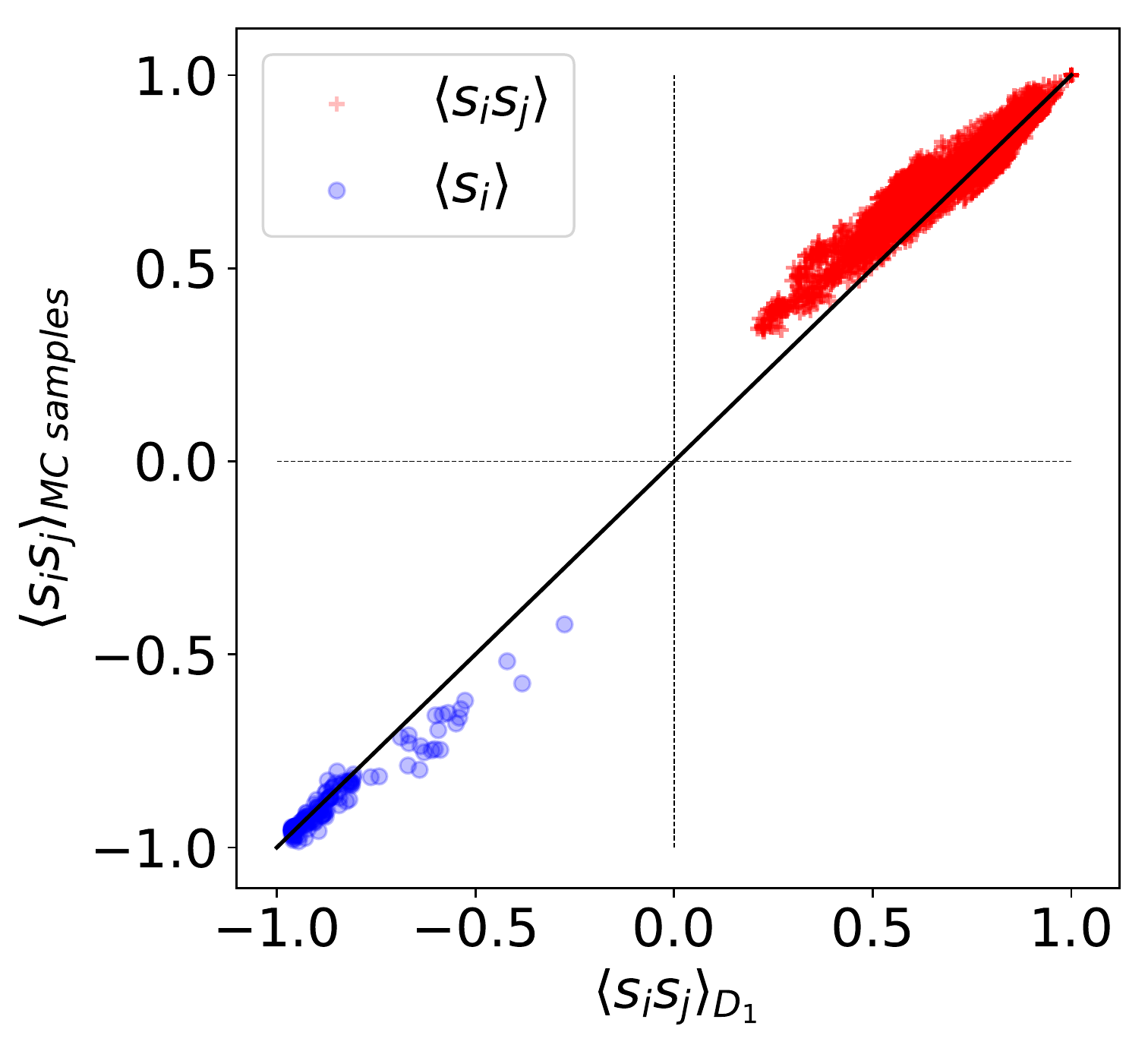}
		\caption{}
		\label{figs: ss-dynamics-D1}
	\end{subfigure}\hfill
	\begin{subfigure}[b]{0.25\textwidth}
		\centering
		\includegraphics[width=.85\linewidth]{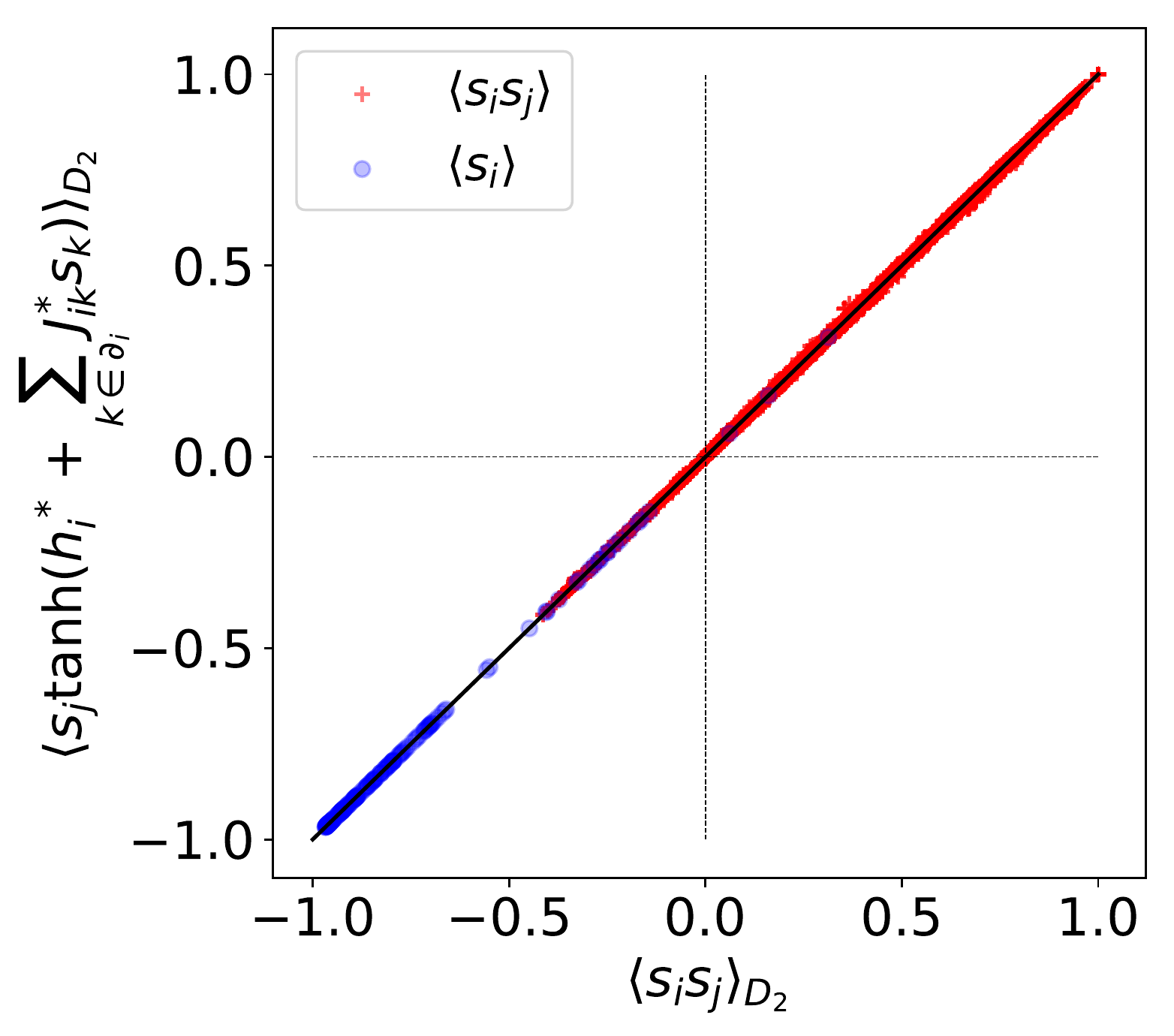}
		\caption{}
		\label{figs: ss-goodness-D2}
	\end{subfigure}%
	\begin{subfigure}[b]{0.25\textwidth}  
		\centering 
		\includegraphics[width=.85\linewidth]{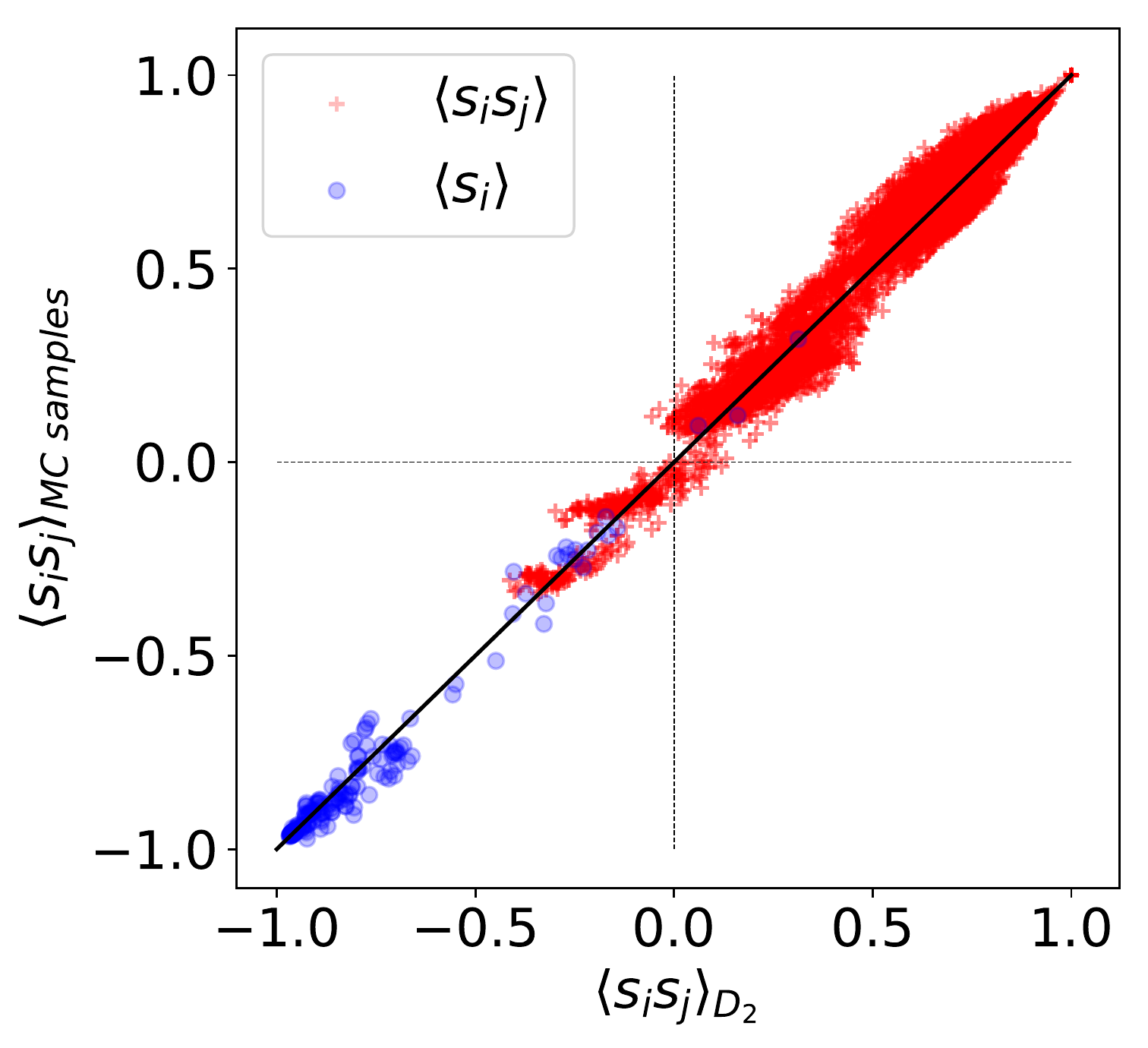}
		\caption{}
		\label{figs: ss-dynamics-D2}
	\end{subfigure}
	
	\caption{Estimated $\avg{s_i}$ and $\avg{s_i s_j}$ against the actual values from data (a,c) reconstructed by applying the learned parameters on the data set $D_1$ and $D2$, and (b,d) using the Monte Carlo samples drawn from Glauber dynamics for data sets $D_1$ and $D_2$.}
	\label{figs: ss-metric}
\end{figure}

Figs. \ref{figs: ss-goodness-D1} and Figs. \ref{figs: ss-goodness-D2} show goodness of fit for the estimated $\avg{s_i}$ and $\avg{s_i s_j}$ reconstructed from Equ.\eqref{equ: ss-derivative} against the values computed from the  corresponding data set where $\avg{s_i} = \avg{s_i r_0}$ with $r_0=1$. The figures show that the learned models fits to the corresponding data. Also, we notice that using data set $D1$ we observe only positive $\avg{s_i s_j}$ for pairwise interactions. However, in data set $D_2$, we have pairs of links with $\avg{s_i s_j} \leq 0$ which means we have lines $i$ and $j$ with $s_i = -s_j$, i.e., only one of them fails in steady state. This observation is the effect of indirect interactions in severe cascading scenarios which is not observed in the normal operation of a power system. Its physical meaning shows that the network partitions in cascading scenarios.   

Next, we generate $M$ samples using the Glauber dynamics starting from a random initial $\V s(0)$ in which each state sets uniformly $+1$ or $-1$. Therefore, the initial network states are very far from the typical steady states used in the training phase, and we need many updates in the Glauber dynamics.    
We set the warm-up time to $10^3 L$ in Monte Carlo simulations and the Monte Carlo step to $20L$ between sampling. Fig. \ref{figs: ss-dynamics-D1} and Fig. \ref{figs: ss-dynamics-D2}, show the $\avg{s_i}$ and $\avg{s_i s_j}$ from these samples against the values in corresponding data sets. Our extensive numerical study shows that the reconstruction of weak (near zero) and negative $\avg{s_i s_j}$ from the Monte Carlo samples is very hard and corresponds to sampling rare events from a dynamical system. This observation also emphasizes that relying on just positive correlations between the line failure is insufficient to understand the system's behavior in large cascades.

\begin{figure}
	\begin{subfigure}[b]{0.30\textwidth}
		\centering
		\includegraphics[width=.85\linewidth]{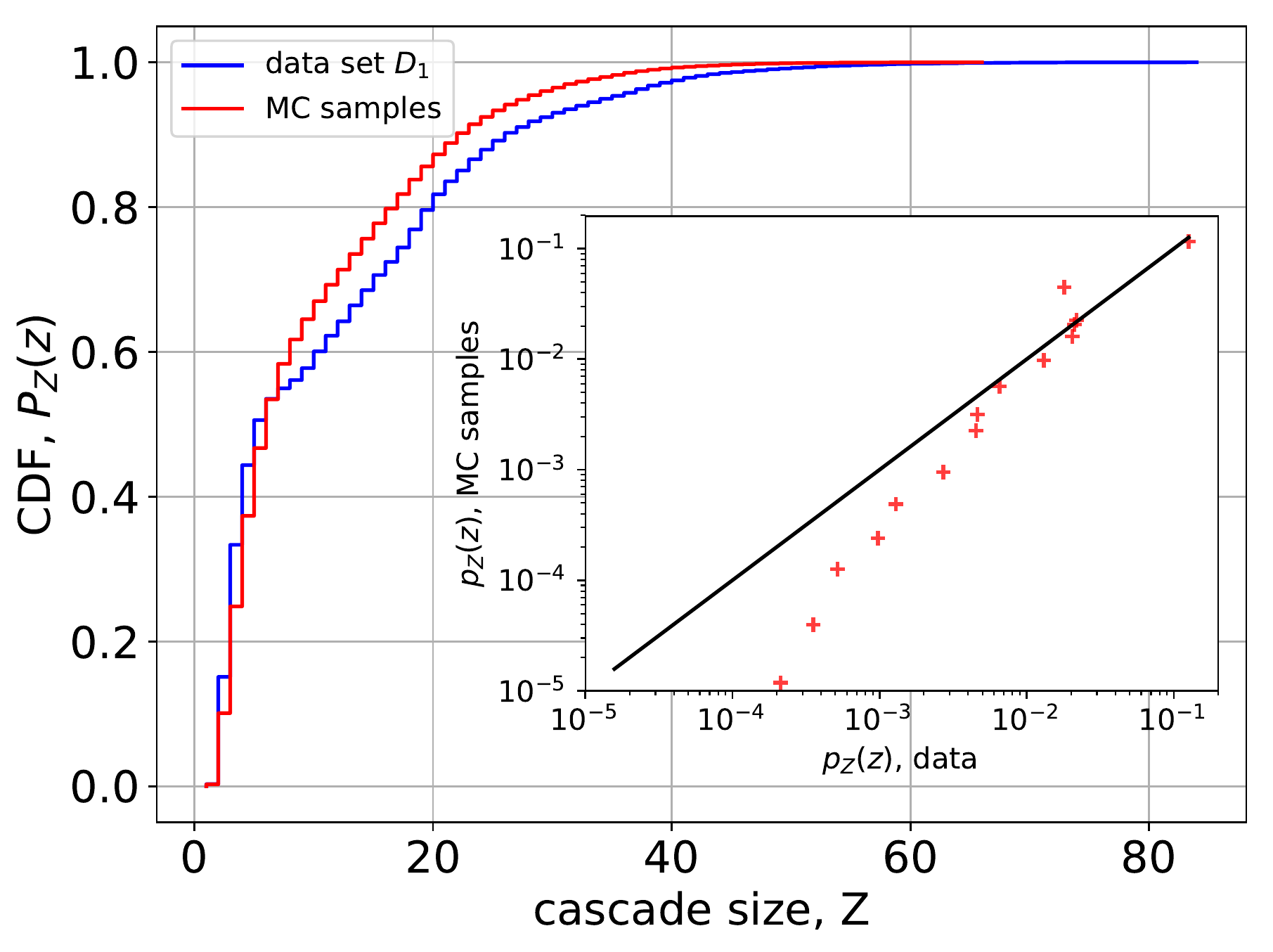}
		\caption{}
		\label{figs: ss-cassize-D1}
	\end{subfigure}%
	\begin{subfigure}[b]{0.20\textwidth}  
		\centering 
		\includegraphics[width=.99\linewidth]{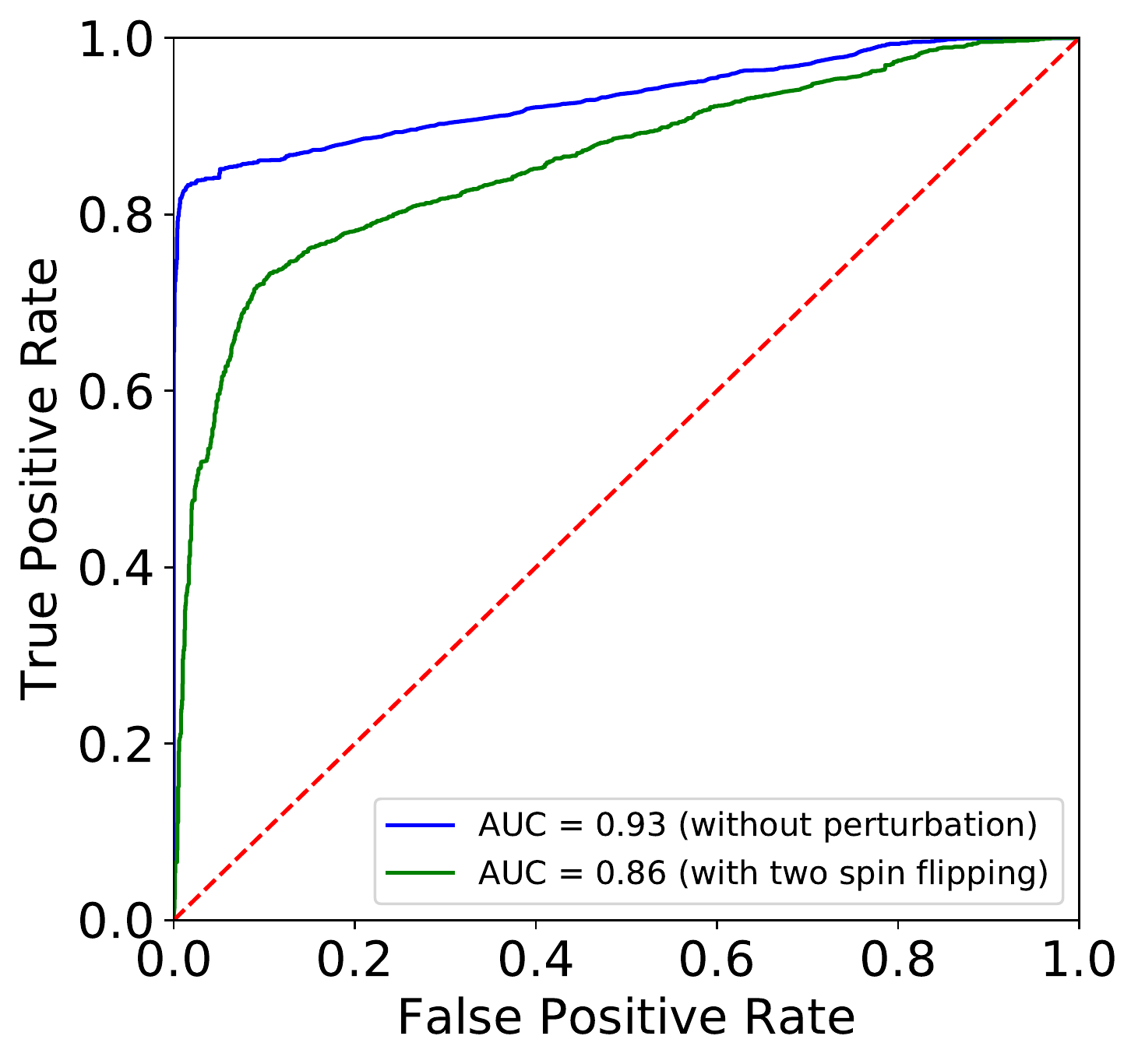}
		\caption{}
		\label{figs: ss-roc-D1}
	\end{subfigure}\hfill
	\begin{subfigure}[b]{0.3\textwidth}
		\centering
		\includegraphics[width=.85\linewidth]{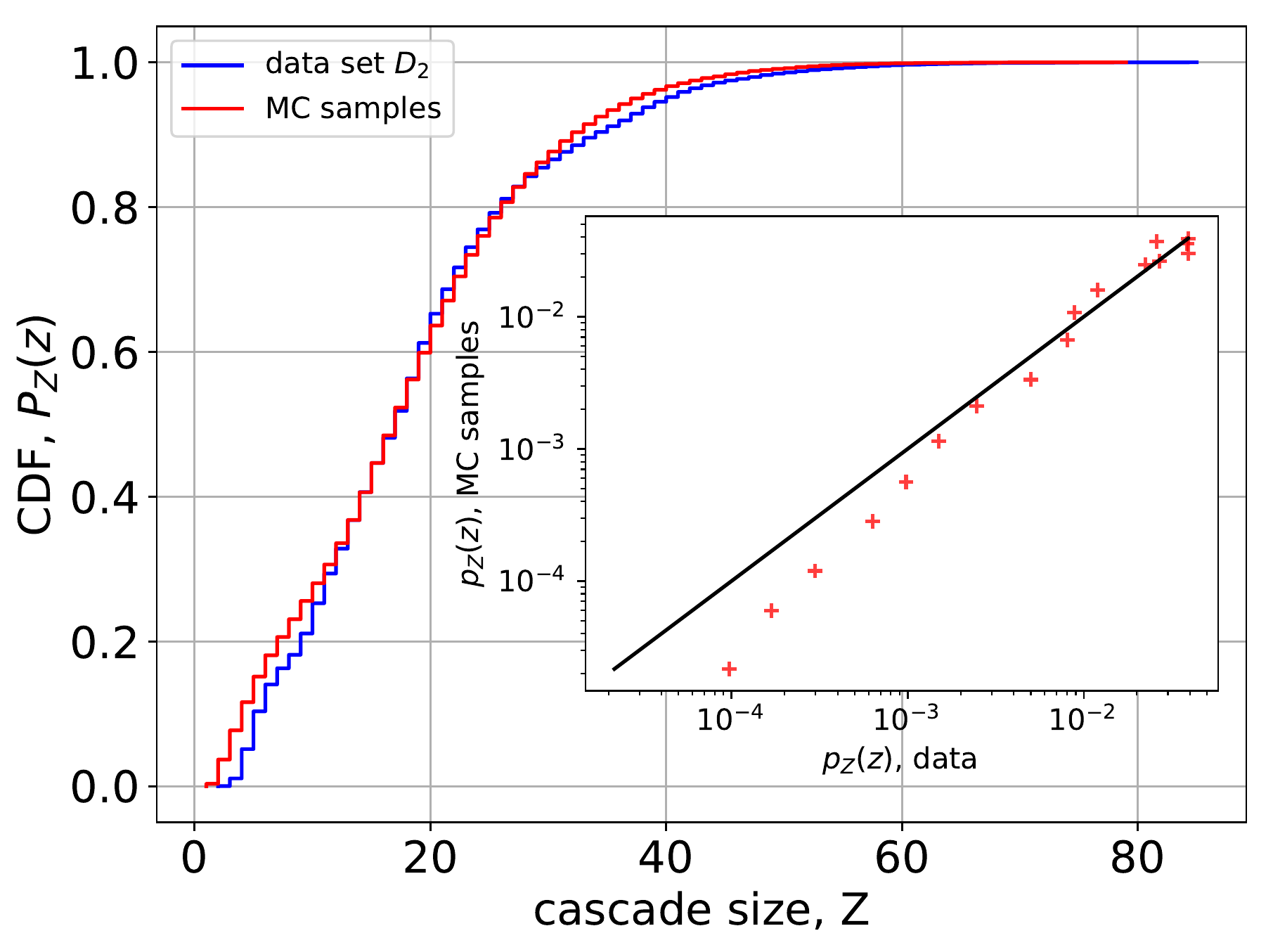}
		\caption{}
		\label{figs: ss-cassize-D2}
	\end{subfigure}%
	\begin{subfigure}[b]{0.2\textwidth}  
		\centering 
		\includegraphics[width=.99\linewidth]{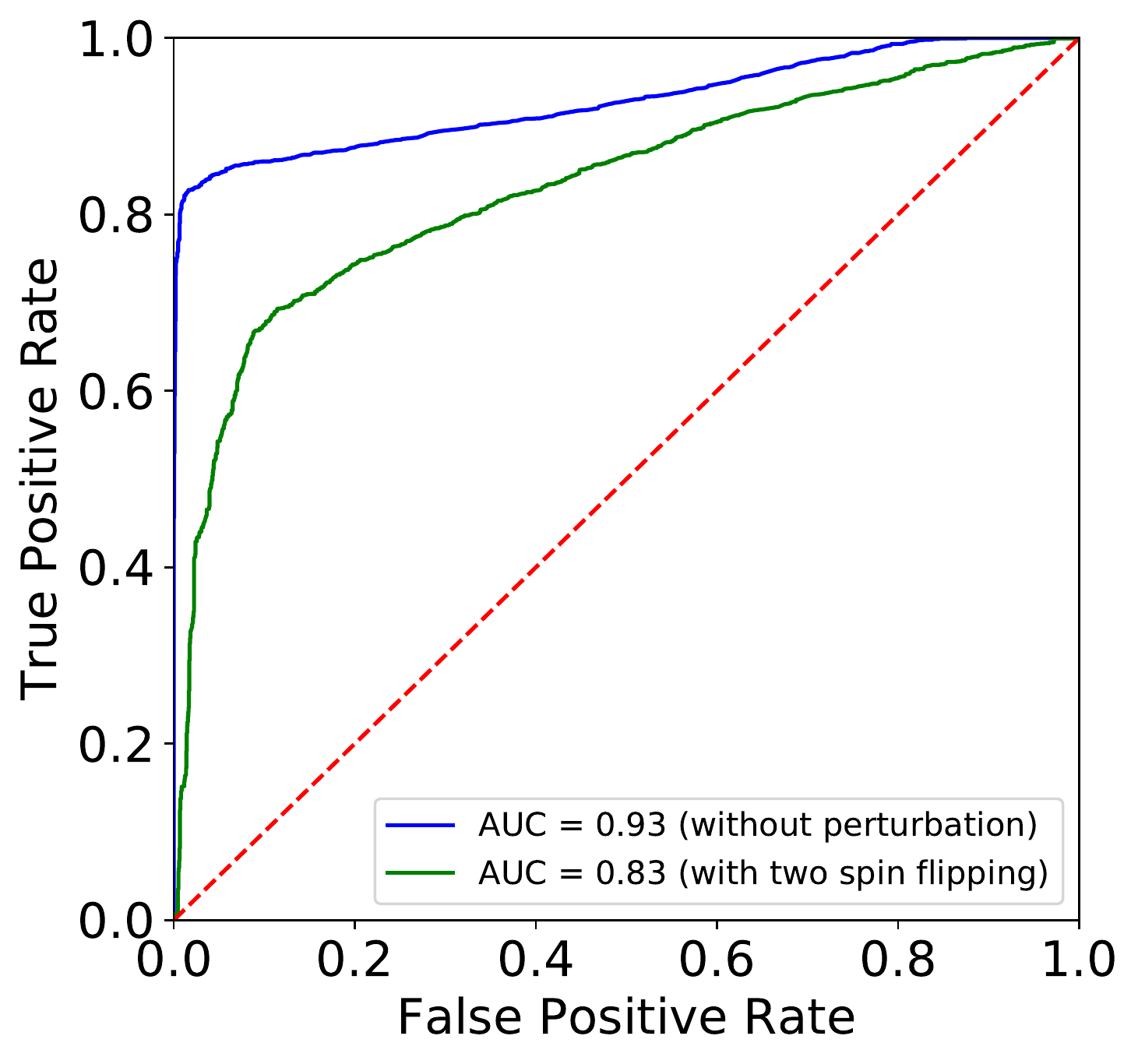}
		\caption{}
		\label{figs: ss-roc-D2}
	\end{subfigure}
	
	\caption{(a,c) CDF of the cascade size from the data sets and the MC samples, the inset compares the binned probability of the cascade size for the MC samples against the values in the corresponding data set. (b,d) The ROC for predicting the state of a selected link without and with two neighbor links state flipping.}
	\label{figs: ss-predict}
\end{figure}

To evaluate the predictive capability of the model, we next compare the cumulative distribution function (CDF) of cascade size, $P_Z$,  for steady state configurations in the Monte Carlo (MC) samples against the data in Fig.~\ref{figs: ss-cassize-D1} and Fig.~\ref{figs: ss-cassize-D2}. The maximum cascade size, the maximum number of failed links, in the data sets are $Z_{D_1}^{max}=84$ and $Z_{D_2}^{max}=85$ and in the MC samples are $Z_{MC_1}^{max}=66$ and $Z_{MC_2}^{max}=79$. As expected, the model learned with more extreme samples better captures the link states in the cascading scenarios. The inset of the figures compares binned probability of the cascade size in which we plot $p_{Z}(z) = \Pr(z \leq Z \leq z+\Delta z)$ with $\Delta z=\frac{Z_{D}^{max}}{20}$ for the MC samples against the values in the corresponding data set. We note that the density function of cascade size, $p_Z(z)$ spans three orders of magnitude, indicating the power-law distribution at the tail. Also, the model successfully generates samples whose density function spans this range.

In another predictive experiment, we generate new $5000$ failure trajectories independently of the training data sets and evaluate how the learned model predicts the state of a specific link given the others' states. For each new sample, we select a link with state $+1$ or $-1$ with the probability of $0.5$. We then predict the selected link's true state probability using the model, assuming that the other links' states are available. Also, we perform the same experiment when we randomly select two neighboring links of the selected link and intentionally flipping their states. Fig. \ref{figs: ss-roc-D1} and Fig. \ref{figs: ss-roc-D2} show the corresponding Receiver Operating Characteristics (ROC) curves for data sets $D_1$ and $D_2$. The ROC curve shows the predictor's performance by depicting the true positive rate against the false positive rate for different thresholds. The models fairly predict the true failure probability of the selected links. The decrease in the ROC's AUC (area under the curve) with perturbations shows the model's sensitivity to perturbing explanatory variables.

\subsection{Inference using interaction matrix}

\begin{figure}[ht!]
	\includegraphics[scale=0.22]{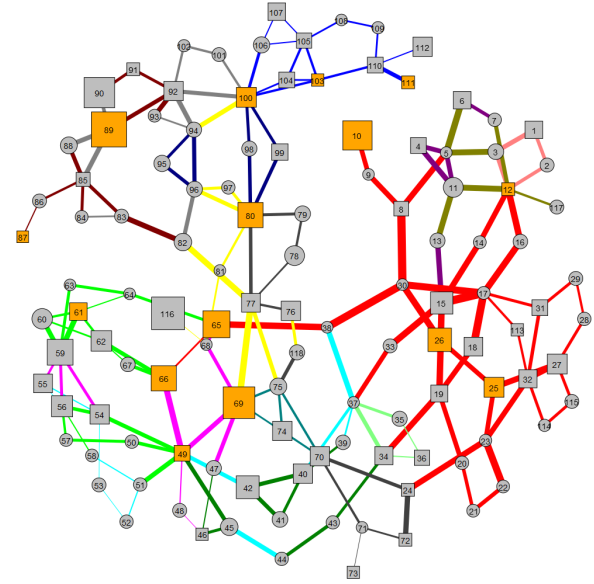}
	\caption{The IEEE-118 network graph where the width of each line reflects the number of other lines which are influenced according to the interaction matrix learned from data set $D2$. Generator and load buses are depicted by rectangles and circles, respectively. The orange and gray colors show the net power generation or consumption at the corresponding node. The size of the node reflects the amount of the net power generation/demand. Lines with the same color are clusters found by Infomap.}\label{figs: 118-interaction}
\end{figure}

In this section, we use the static interaction matrix to infer some structural properties of the network. Fig.~\ref{figs: 118-interaction} shows the connectivity graph of the IEEE-118 bus network in which the width of each line reflects the influential impact of the line according to the learned $\M J$ matrix for data set $D_2$, i.e., the number of other lines which are affected by the state of this line. As intuitively expected, the most influential lines are connected to big generation points (large orange rectangles), and the least ones connect small loads (small grey circles) to the network.

We next study the regularities in the interaction graph, $\tilde{\mathcal{G}}$, which corresponds to the interaction matrix $\M J$ to find links that fail together. $\tilde{\mathcal{G}}$ is weighted, signed, and directed graph with $L$ nodes in which a link $i \rightarrow j$ shows that line $i$ affects the state of the line $j$.

We are interested in finding co-susceptible groups of lines that tend to fail together statistically. We use the Infomap \cite{Rosvall1118} as an appropriate algorithm with proper weights for each interaction to find clusters of nodes with the same states in different network steady states. Infomap is a flow-based clustering mechanism that finds the organization based on the real flow of interactions in the underlying network. Here, we use Infomap to capture the desired failure propagation dynamics (flow) in our directed, and weighted interaction graph \cite{rosvall2009map}.

We first convert the interaction values to proper positive weights, which the random walker subsequently uses in the network as a proxy of failure flow in the network. Let $p_i~=~\Pr(s_i=+1 \given \V s_{\partial_i})$ where we remove time dependency for short writing. In the binary logistic regression learning we find $(h_i^*, \M J_i^*)$ such that $\log \frac{p_i}{1 - p_i} ~=~ 2(h_i^* + \sum_{j \in{\partial i}}J_{ij}^*s_j)$, i.e., we find the log-odds of line $i$ failure in terms of the explanatory neighboring links' states. Now, assume the random walker is at node $j \in \partial_i$ of $\tilde{\mathcal{G}}$. The state of node $j$ contributes in node $i$' state according to $[\M J]_{ij}$.  Let $p_{ij}^+ = \Pr(s_i = +1 \given s_j=+1, \V s_{\partial_i \setminus j})$ and $p_{ij}^- = \Pr(s_i = +1 \given s_j=-1, \V s_{\partial_i \setminus j})$. Using \eqref{equ: ss-prob} we observe that \cite{bresler2017learning}

\begin{align}
	e^{4J_{ij}} = \frac{p_{ij}^+(1-p_{ij}^-)}{p_{ij}^-(1-p_{ij}^+)}. \label{equ:info-weight}
\end{align}


We can interpret $p_{ij}^+$ as the probability of failure flow from $j$ to $i$ for a given $\V s_{\partial_i \setminus j}$ where $\frac{p_{ij}^+}{1 - p_{ij}^+}$ is the corresponding odds. Correspondingly, $p_{ij}^-$ is the probability of failure flow from $i$'s neighbors except $j$ to $i$. The ratio $\big[p_{ij}^+/(1-p_{ij}^+)\big]/\big[p_{ij}^-/(1-p_{ij}^-)\big]$ is a good measure for the share of failure flow from $j$ to $i$. Therefore, we assign $e^{4J_{ij}}$ as the weight of link $j \rightarrow i$ in $\tilde{\mathcal{G}}$.

If $J_{ij}$ is sufficiently positive, then $p_{ij}^+ \gg p_{ij}^-$ and if $J_{ij}$ is sufficiently negative $p_{ij}^+ \ll p_{ij}^-$. Note that weak coupling $J_{ij} \approx 0$ means $p_{ij}^+ = p_{ij}^-$ and as expected does not contribute much in clustering process. We run the two-level Infomap clustering algorithm on data sets $D_1$ and $D_2$ and sort the clusters based on their sizes. The nodes of $\tilde{\mathcal{G}}$ (lines of $\mathcal{G}$) belong to the same cluster, then get sequential indices.

\begin{figure}
	\centering
	\begin{subfigure}[b]{0.25\textwidth}
		\centering
		\includegraphics[width=.95\linewidth]{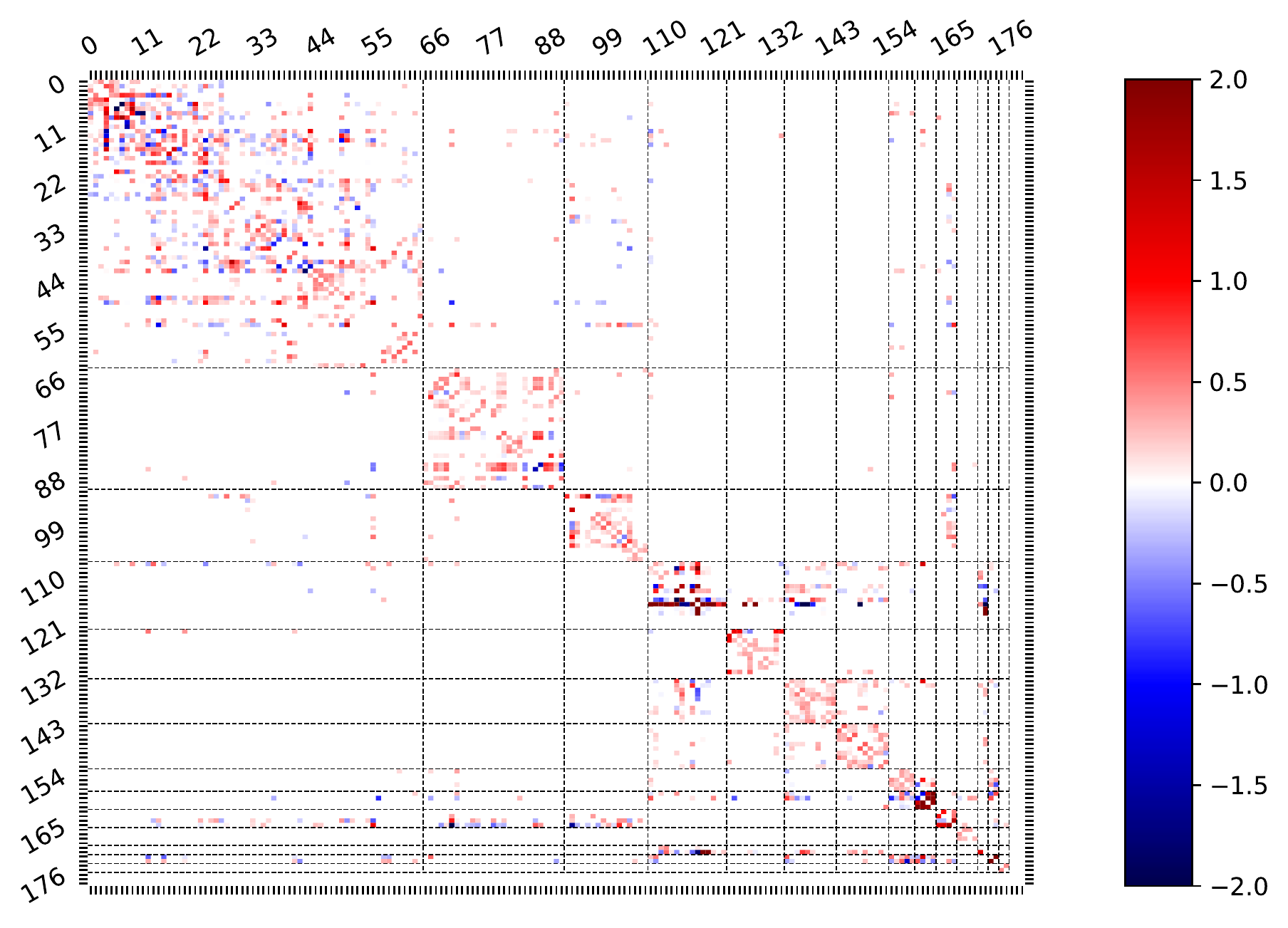}
		\caption{}    
		\label{figs: ss-infomap-D1}
	\end{subfigure}%
	\begin{subfigure}[b]{0.25\textwidth}
		\centering
		\includegraphics[width=.95\linewidth]{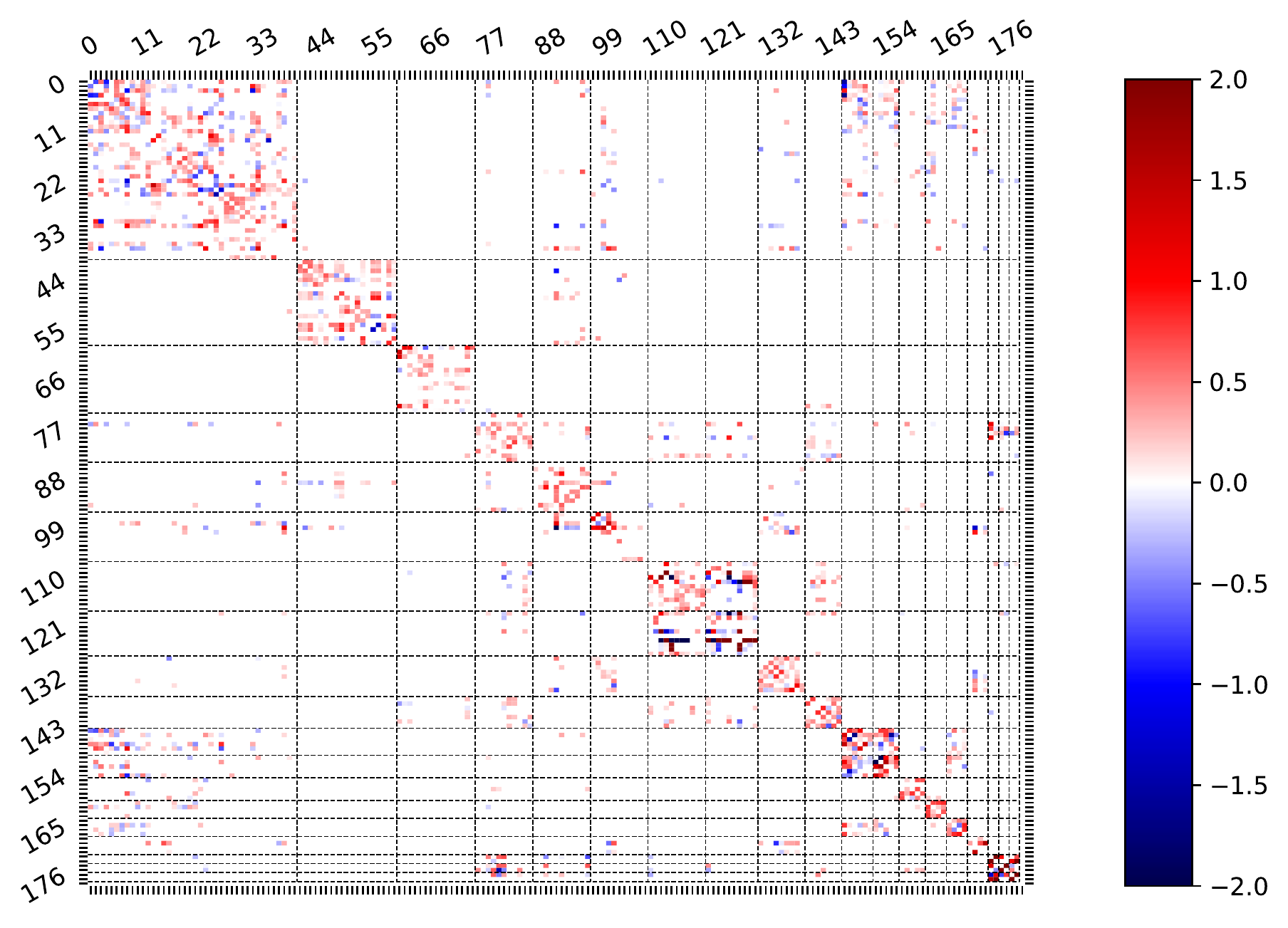}
		\caption{}    
		\label{figs: ss-infomap-D1}
	\end{subfigure}
	\begin{subfigure}[b]{0.25\textwidth}  
		\centering 
		\includegraphics[width=.95\linewidth]{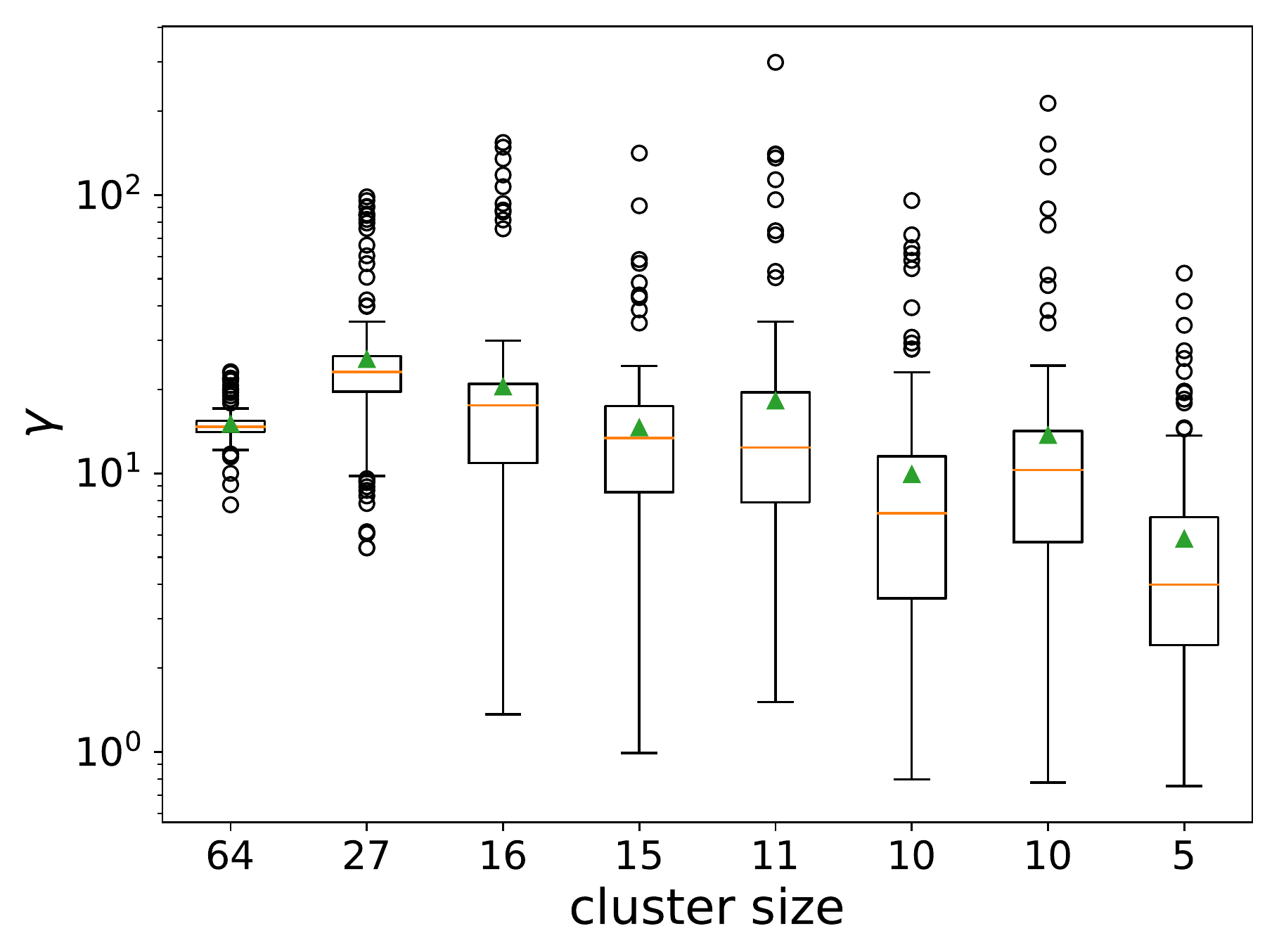}
		\caption{}    
		\label{figs: ss-gamma-D1}
	\end{subfigure}%
	\begin{subfigure}[b]{0.25\textwidth}
		\centering
		\includegraphics[width=.95\linewidth]{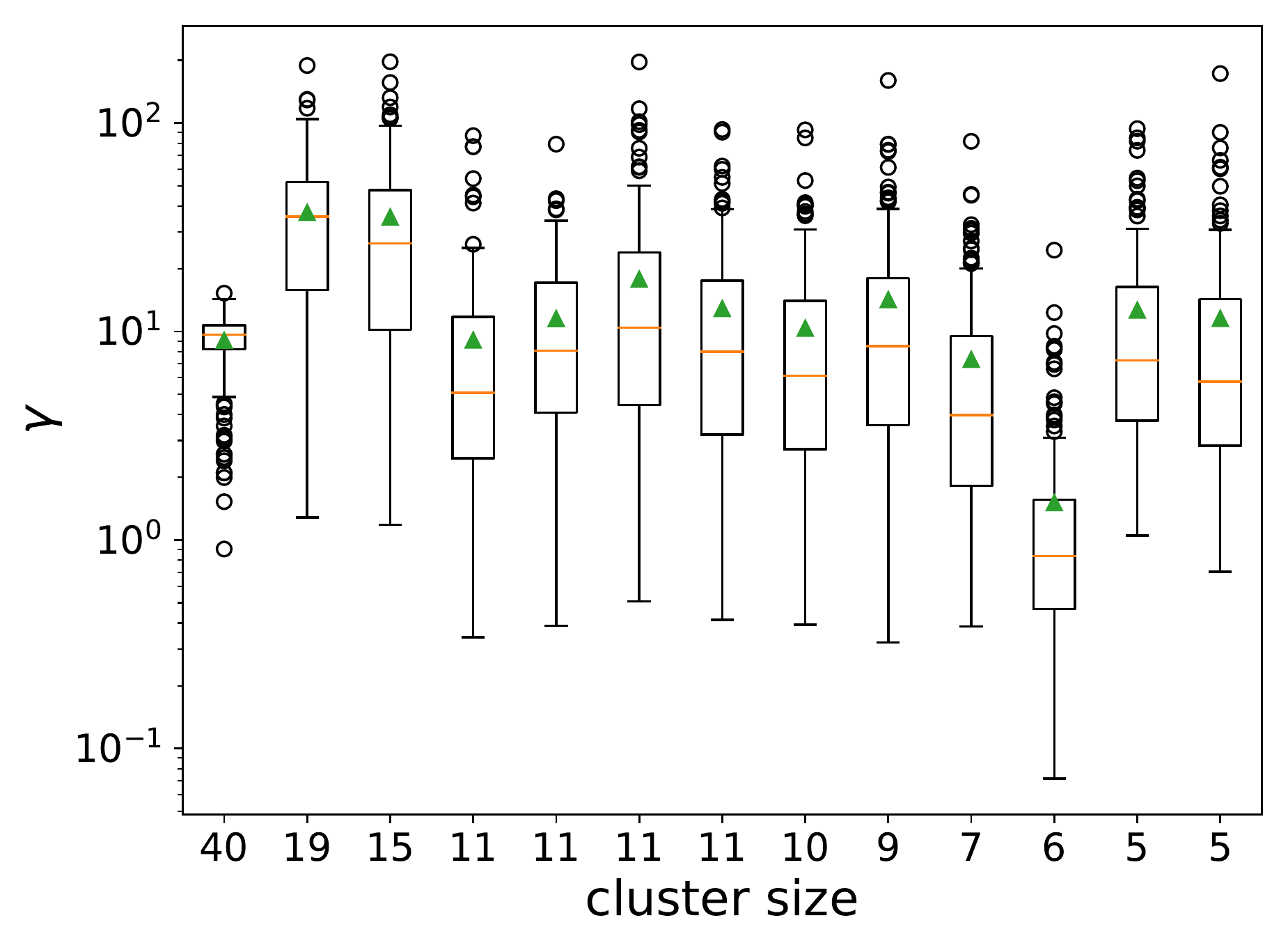}
		\caption{}    
		\label{figs: ss-gamma-D2}
	\end{subfigure}
	
	\caption{(a,b)-heat map of the interaction matrix when lines are grouped and reindexed sequentially based on the Infomap clustering of the corresponding interaction graph $\mathcal{G}$ for (a) data set $D_1$ (b) data set $D_2$. the thin dashed lines separate different clusters. (c,d)-the box plot of the $\gamma$ values where the triangle token shows the mean and the horizontal line in each box shows the median for (c) data set $D_1$ and (d) data set $D_2$}
	\label{figs: ss-infomaps}
\end{figure}

Fig.~\ref{figs: ss-infomaps} shows the results for both data sets where we sort clusters according to their sizes and assign consecutive indices to lines in the same clusters. Infomap finds 8, and 15 clusters with cluster sizes greater than two for $D_1$ and $D_2$. The models suggest that there exists a clustering structure in the line failure in both data sets. In Fig.~\ref{figs: 118-interaction}, the lines which are grouped in the same cluster by the Infomap mechanism for $D_2$  have the same color. 
As expected, the nearby lines are mostly in the same cluster. We, however, observe distant lines which are grouped in the same cluster. Furthermore, the clustering result for data set $D_2$ shows more distinctive clusters roots to line pairs with $\avg {s_i s_j} \approx 0$.

Let random variable $Z_{\mathcal{C}}=\sum_{j \in \mathcal{C}} \frac{(1+s_j)}{2}$ denote the number of failures in a final steady-state cascading trajectory for cluster $\mathcal{C}$. We compute $\Pr({Z_{\mathcal{C}} = z \given Z_{\mathcal{C}} > 0})$ by marginalizing over the other lines' states in the data set to find to what extent the failure of one line in the group leads to other lines' failures in this group. The null hypothesis is to select a subset of lines randomly and uniformly, $\mathcal{R}$, with the same cardinality, i.e.,   $|\mathcal{C}| = |\mathcal{R}|$, and compute the same measure. The ratio of
$\gamma = \frac{\E{[Z_{\mathcal{C}} = z \given Z_{\mathcal{C}} > 0]}}{\E{[Z_{\mathcal{R}} = z \given Z_{\mathcal{R}} > 0]}}$ then shows the effectiveness of the clustering method against the null hypothesis. Here $\E$ denotes the expectation value of the desired co-failure measure. Fig. (\ref{figs: ss-gamma-D1}) and Fig. (\ref{figs: ss-gamma-D2}) show the distribution of the $\gamma$ values for 200 random samples as a box plot chart for cluster sizes greater than four where the triangle token shows the mean and the horizontal bar in each box is the median of samples.  We observe that except for one cluster in data set $D_2$, the mean values of the co-susceptibility measure $\gamma$ in the Infomap clusters are approximately one order of magnitude greater than the null hypothesis.

\section{Time series interaction modeling} \label{sec: ts-modeling}
The objective of this section is to learn how the states of links change over time. Instead of updating a specific link-state near the steady states, we find the interaction matrix that encodes how the cascade unfolds in the network. The importance of this problem is on designing mitigation strategies for power networks.

Each trajectory in our data sets captures the sequence of all link states until the network settles in a steady state. Therefore, each trajectory is a time series of links' states $\V s(0), \V s(1), \ldots \V s(t_{ss})$ where $t_{ss}$ is the time that failure propagation ends.
The next state of the steady state is itself, $\V s(t_{ss}+1) = \V s(t_{ss})$. For each data set we remove possible duplicate trajectories due to the same initial failure, and find $T = \sum_{j=1}^{M_0} 1+ t_{ss}^j $ consecutive network's state.

\subsection{Logistic regression model}
We adopt the kinetic Ising model with asynchronous updates \cite{zeng2013maximum}. In this model, at each time step the state of each link is updated with the probability given in \eqref{equ: ss-prob} which can be read as $\Pr(s_i(t+1) | \V s_{-i}(t)) = \frac{e^{s_i(t+1)H_i(t)}}{2 \cosh{H_i(t)}}$. Note that the deployed model and data sets of steady states in the previous section can be considered as one step kinetic Ising model. The likelihood function is 

\begin{align}
	\mathcal{L}_D (\V h, \M J) = \frac{1}{T} \sum_{t=1}^{T-1} \sum_{i=1}^{L} \Big[s_i(t+1)H_i(t) - \ln2\cosh(H_i(t))\Big]. \label{figs: ts-likelihood} 
\end{align}

The objective is finding $(\V h, \M J)$ which maximize the desired $l_1$-regularized function
\begin{align}
	(\V h^*, \M J^*) = \textrm{argmax}_{(\V h, \M J)} \mathcal{L}_D(\V h, \M J) - \lambda \sum_{j \neq i}|d_{ij} J_{ij}|,\label{equ: ts-objective} 
\end{align}

In contrast to the previous section in which we solve an optimization problem for each link independently, we should find $(\V h, \M J)$ in an optimization problem over $L^2$ variables. Likewise, we follow a two-stage algorithm in the previous section to find the most explanatory interactions and fine-tune them.  Since these are convex optimization problems, there are very efficient numerical methods to solve these problems. We use the naive gradient descent method to find the solution. 

Computing the derivative of the likelihood function we have:

\begin{align}
	\frac{\partial \mathcal{L}}{\partial h_i} &= \frac{1}{T}\sum_{t=0}^{T-1}\big[s_i(t+1) - \tanh(H_i(t))\big] \\
	\frac{\partial \mathcal{L}}{\partial J_{ij}} &= \frac{1}{T}\sum_{t=0}^{T-1}s_j(t)\Big[s_i(t+1) - \tanh(H_i(t))\Big]. \label{equs: ts-obj-derivative}
\end{align}

Therefore, at the optimal point we have $\avg{s_i(t)}_D^t = \avg{\tanh(H_i(t))}_D^t$ and $\avg{s_i(t) s_j(t+1)}_D^t = \avg{s_j(t)\tanh(H_i(t))}_D^t$ where $\avg{f(s(t))}_D^t = \frac{1}{T}\sum_{t=0}^{T-1} f(s(t))$ which used as a goodness of fit measure.  

\subsection{Time series interactions}
\begin{figure}
	\centering
	\begin{subfigure}[b]{0.25\textwidth}
		\centering
		\includegraphics[width=\textwidth]{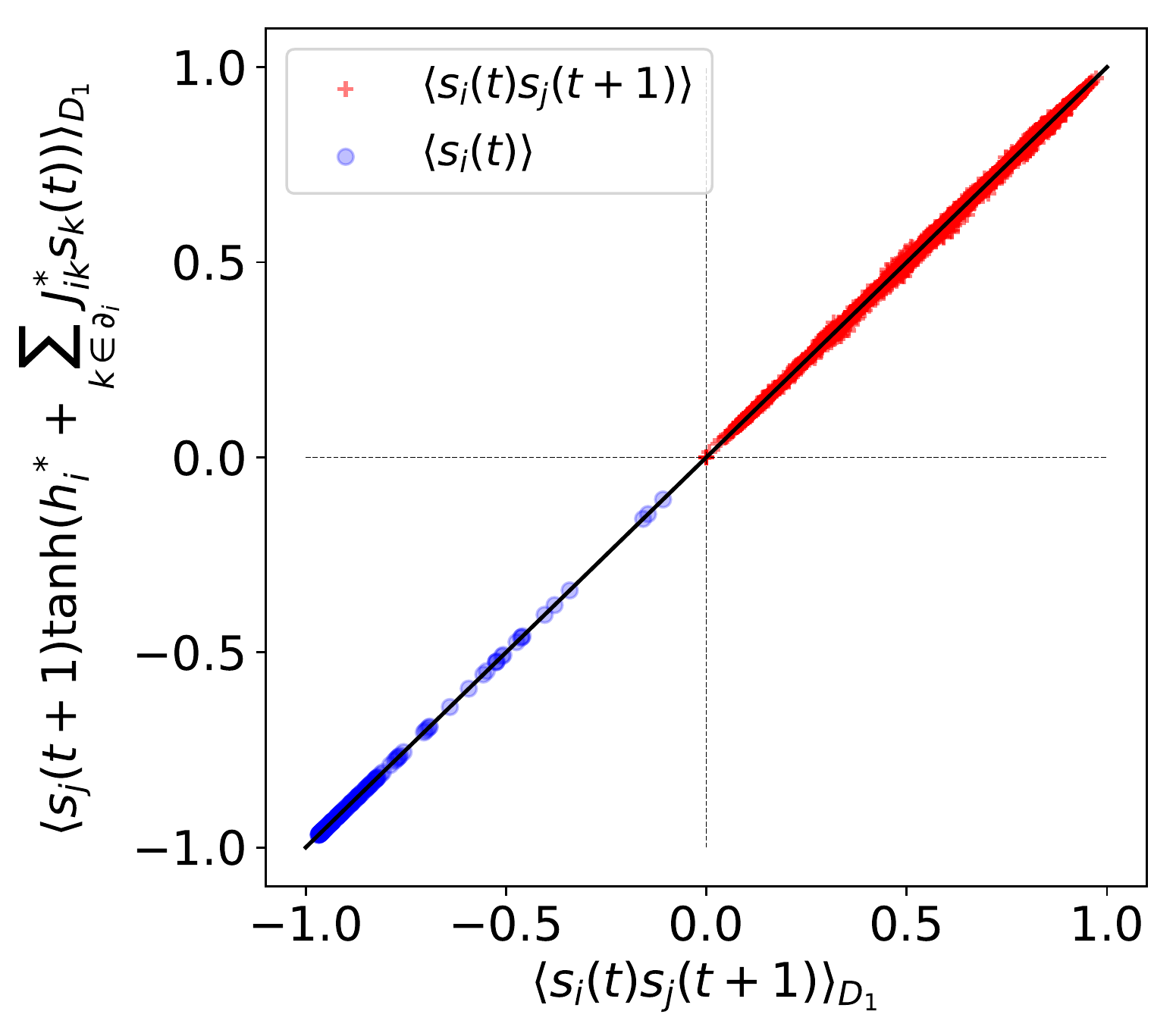}
		\caption{}    
		\label{figs: ts-fitness-D1}
	\end{subfigure}%
	\begin{subfigure}[b]{0.25\textwidth}  
		\centering 
		\includegraphics[width=\textwidth]{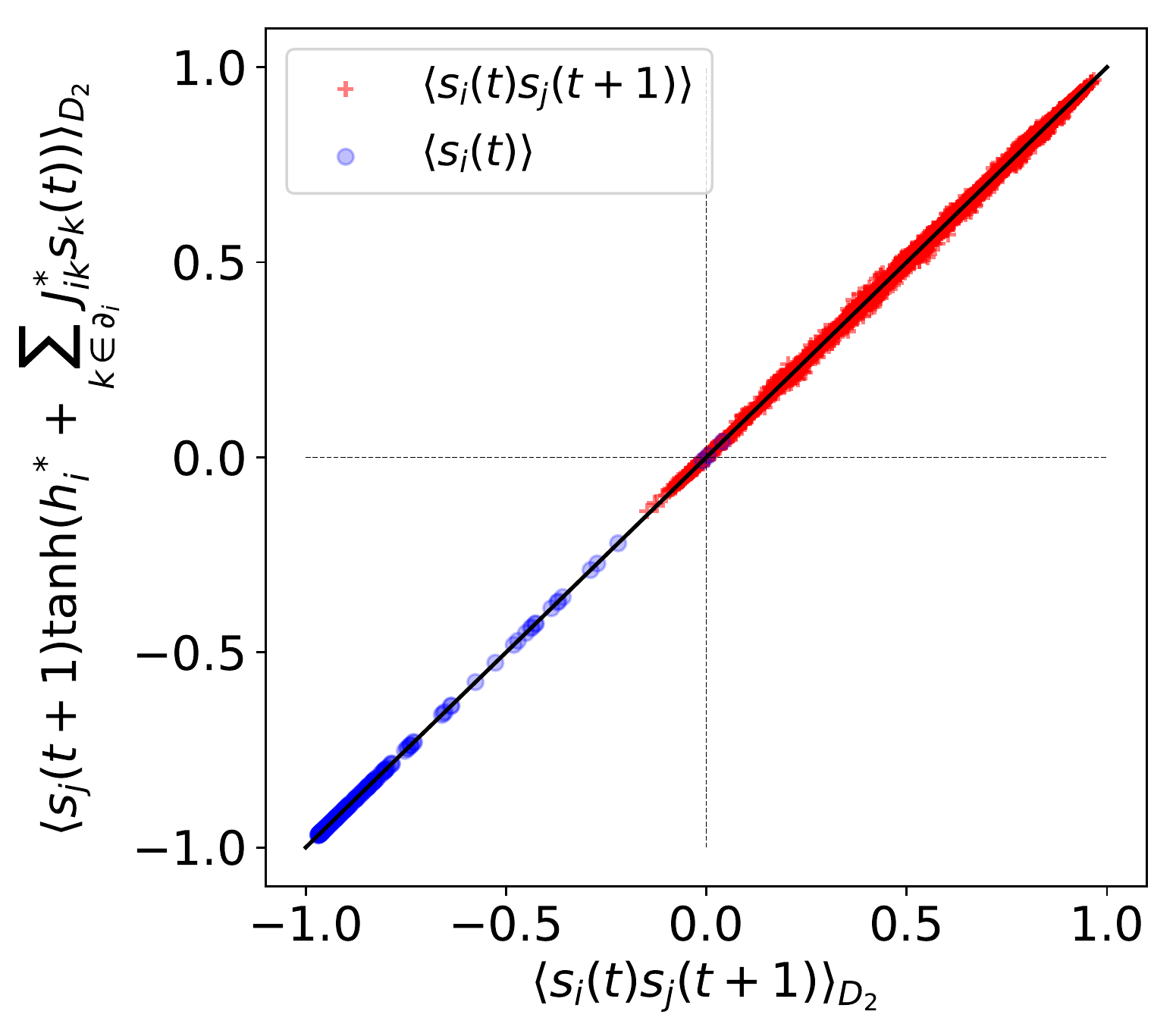}
		\caption{}    
		\label{figs: ts-fitness-D2}
	\end{subfigure}
	
	\caption{Estimated $\avg{s_i(t)}$ and $\avg{s_i(t) s_j(t+1)}$ against the actual values from data sets for (a) data set $D_1$ (b) data set $D_2$.}
	\label{figs: ts-fitness}
\end{figure}

We set the same parameter values for $\lambda$ and $\delta$ as the steady state analysis in order to find the corresponding $(\V h, \M J)$ for each data set. Fig.~\ref{figs: ts-fitness} shows that the model appropriately reconstructs $\avg{s_i(t)}$ and $\avg{s_i(t) s_j(t+1)}$.   

The next step is to measure how the learned model predicts the failure unfolding in time. Here, we should select a threshold for binary decision-making at each step based on each line's predicted probability. We update the network state at each time step and find consecutive network states in the time horizon. The time horizon equals the corresponding trajectory's actual steps before settlement.
Note that the possible prediction error at a time step will propagate to the consecutive time step predictions. 

\begin{figure}
	\centering
	\begin{subfigure}[b]{0.25\textwidth}
		\centering
		\includegraphics[width=\textwidth]{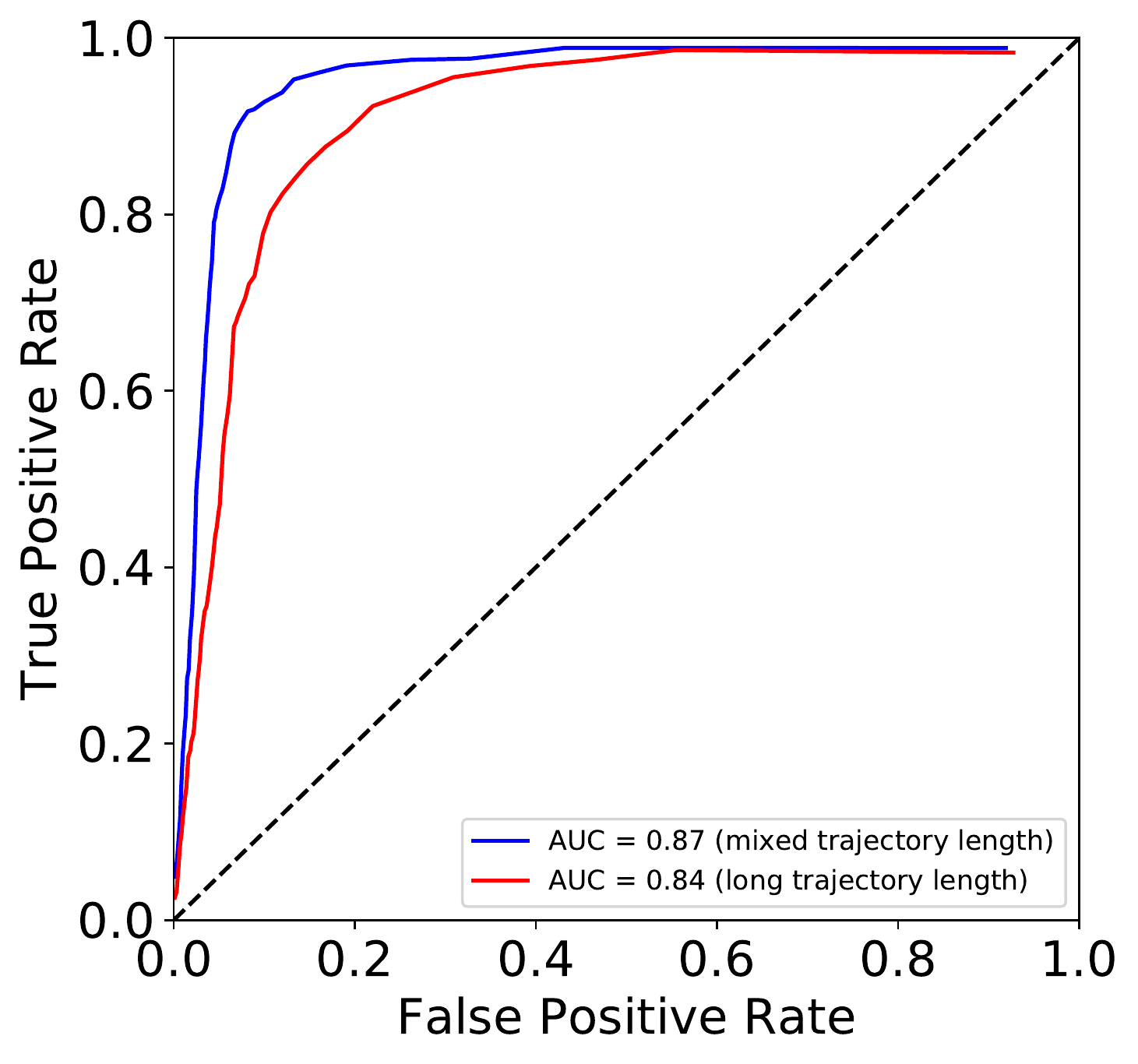}
		\caption{}    
		\label{figs: ts-roc-D2}
	\end{subfigure}%
	\begin{subfigure}[b]{0.25\textwidth}  
		\centering 
		\includegraphics[width=\textwidth]{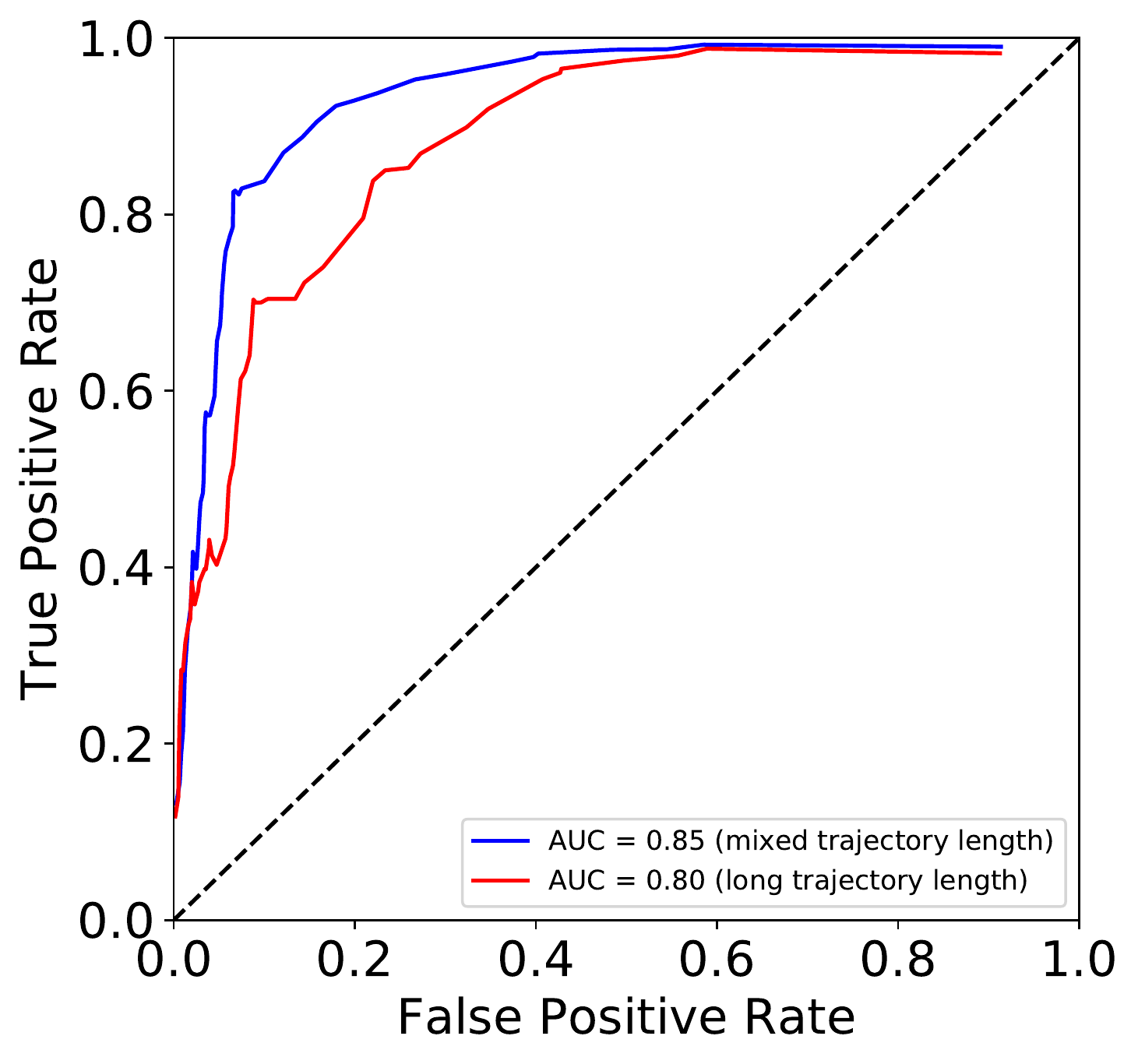}
		\caption{}   
		\label{figs: ts-roc-D2-inf}
	\end{subfigure}
	
	\caption{the ROC curves for predicting the network state in time horizon compared to the ground truth trajectory for data set $D_2$ (a) in the time horizon equal to the actual trajectory (b) until no new updates happen in the network's state}
	\label{figs: ts-roc}
\end{figure}

We find the consecutive network states for different threshold values and compare the predicted set of failed lines against the ground truth for 1000 independent trajectories of failure cascading. We compute the corresponding true positive and false-positive rates and find the ROC curve. Here false positive is predicting a line failure against the ground truth. See Fig.~\ref{figs: ts-roc-D2}.
We repeat this experiment over another 1000 trajectories that last at least six-time steps to see how well the consecutive line failure prediction works—the corresponding ROC curve named as long-trajectories in  Fig.~\ref{figs: ts-roc-D2}. We find similar results for data set $D_1$.

Finally, we repeat this experiment in the time horizon until no update happens in the network's state. Fig.~\ref{figs: ts-roc-D2-inf} shows the corresponding ROC curve. These results show that the learned dynamic interaction matrix successfully predicts the network's state in consecutive time steps until settlement at the final steady-state.

\section{Conclusion and future works} \label{sec: conclusion}
We find static and dynamic interaction models from steady-state and trajectories of consecutive line failures in a power grid network.
We use weighted regularized regression-based machine learning techniques to find the corresponding model interaction matrix.  
The static model uses conditional maximum entropy learning of predicting each line's state given the states of others near a steady network state.
This model helps find the co-susceptible group of lines that tend to fail together, considering possible indirect interactions. The dynamic model predicts the temporal unfolding of initial failure over time. The results show that the machine learning-based techniques can capture the latent indirect higher-order interactions in failure cascading. Both conditional and time-series learning-based models enjoy the causal representation of the data. Causal learning and inference have recently found many applications in diverse fields and can also help predict and mitigate extreme events in network-based systems. In future works, we analyze the properties of the learned interaction matrices and their relations to the cascading process in power networks.    

%

\section*{Acknowledgments}

A. Ghasemi gratefully acknowledges support from the Max Planck Institute for the Physics of Complex Systems and from the Alexander von Humboldt Foundation for his visiting research in Germany.

\bibliographystyle{IEEEtran} 
\bibliography{refs}

\begin{IEEEbiography}[{\includegraphics[width=1in,height=1.25in,clip,keepaspectratio]{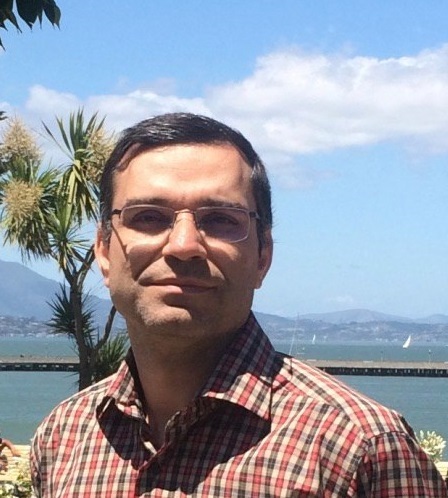}}]{Abdorasoul Ghasemi}
is with the Faculty of Computer Engineering of K.N. Toosi University of Technology, Tehran, Iran. He received his Ph.D. and M.Sc. degrees from Amirkabir University of Technology (Tehran Polytechnique), Tehran, Iran, and his B.Sc. from the Isfahan University of Technology, all in Electrical Engineering. He has spent sabbatical leave with the computer science department at the University of California, Davis, CA, the USA (April 2017 to August 2018) and Max Planck Institute for the physics of complex systems, Dresden, Germany (Dec. 2020 to July 2021). He has been awarded the Alexander von Humboldt fellowship for experienced researchers in July 2021, working on resilient cyber-physical energy systems at the University of Passau, Germany. His research interests include network science and its engineering applications, including communications, energy, and cyber-physical systems using optimization and machine learning approaches. 
\end{IEEEbiography}
\begin{IEEEbiography}[{\includegraphics[width=1in,height=1.25in,clip,keepaspectratio]{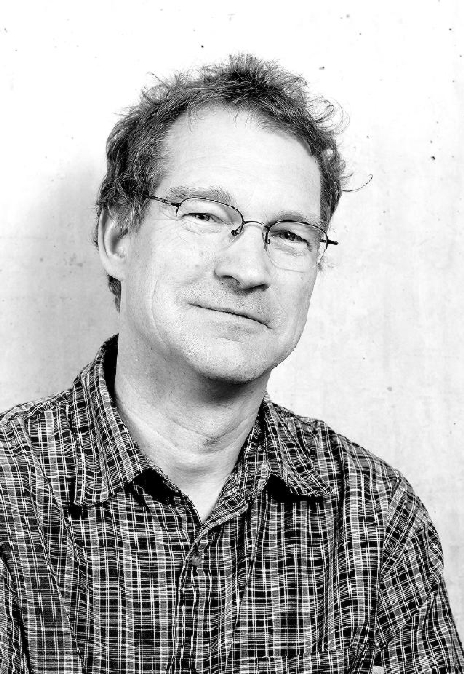}}]{Holger Kantz} 
is head of the Nonlinear Dynamics and Time Series Analysis Research Group at
the Max Planck Institute for the Physics of Complex Systems in Dresden, and an
Adjunct Professor in Statistical Physics at the Technical University of
Dresden. He obtained his Diplom in Physics at the University of Wuppertal in
1986, and completed a PhD in Physics under the supervision of Peter
Grassberger in 1989. After a period as a Postdoctoral Fellow in Florence, he
retuned to Wuppertal in 1991 as a scientific and teaching assistant, and
completed his Habilitation in Theoretical Physics in 1996. Since 1995 he has
been research group leader in Dresden.
His research interests include deterministic chaos, nonlinear stochastic
processes, statistical physics, time series analysis, with applications to
meteorological data and extreme weather events. He has published more than 200 articles in international journals on these topics, as well as 3 volumes of proceedings, and is the co-author (with Thomas Schreiber) of a textbook on Nonlinear Time Series Analysis which is published by Cambridge University Press.
\end{IEEEbiography}

\vfill


\end{document}